\documentclass{article}
\usepackage[latin9]{inputenc}
\usepackage{amsmath}
\usepackage{graphicx}
\usepackage{esint}

\makeatletter
\usepackage[final, nonatbib]{nips_2018}

\usepackage{hyperref}       
\usepackage{url}            
\usepackage{booktabs}       
\usepackage{amsfonts}       
\usepackage{nicefrac}       
\usepackage{microtype}      
\usepackage{amsmath}
\usepackage{graphicx}

\global\long\def\cond#1#2{\left(#1\left|#2\right.\right)}

\title{Statistical mechanics of low-rank tensor decomposition}

\author{
  Jonathan Kadmon \\
  Department of Applied Physics, Stanford University\\
  \texttt{kadmonj@stanford.edu} \\
   \And
   Surya Ganguli \\
   Department of Applied Physics, Stanford University and Google Brain, Mountain View, CA \\
   \texttt{sganguli@stanford.edu} \\
}

\makeatother

\begin{document}

\title{Statistical mechanics of low-rank tensor decomposition }
\maketitle
\begin{abstract}
Often, large, high dimensional datasets collected across multiple
modalities can be organized as a higher order tensor. Low-rank tensor
decomposition then arises as a powerful and widely used tool to discover
simple low dimensional structures underlying such data. However, we
currently lack a theoretical understanding of the algorithmic behavior
of low-rank tensor decompositions. We derive Bayesian approximate
message passing (AMP) algorithms for recovering arbitrarily shaped
low-rank tensors buried within noise, and we employ dynamic mean field
theory to precisely characterize their performance. Our theory reveals
the existence of phase transitions between easy, hard and impossible
inference regimes, and displays an excellent match with simulations.
Moreover, it reveals several qualitative surprises compared to the
behavior of symmetric, cubic tensor decomposition. Finally, we compare
our AMP algorithm to the most commonly used algorithm, alternating
least squares (ALS), and demonstrate that AMP significantly outperforms
ALS in the presence of noise. 
\end{abstract}

\section{Introduction}

The ability to take noisy, complex data structures and decompose them
into smaller, interpretable components in an unsupervised manner is
essential to many fields, from machine learning and signal processing
\cite{anandkumar2014tensor,sidiropoulos2017tensor} to neuroscience
\cite{williams2017unsupervised}. In datasets that can be organized
as an order $2$ data matrix, many popular unsupervised structure
discovery algorithms, like PCA, ICA, SVD or other spectral methods,
can be unified under rubric of low rank matrix decomposition. More
complex data consisting of measurements across multiple modalities
can be organized as higher dimensional data arrays, or higher order
tensors. Often, one can find simple structures in such data by approximating
the data tensor as a sum of rank $1$ tensors. Such decompositions
are known by the name of rank-decomposition, CANDECOMP/PARAFAC or
CP decomposition (see \cite{kolda2009tensor} for an extensive review).

The most widely used algorithm to perform rank decomposition is alternating
least squares (ALS) \cite{carroll1970analysis,harshman1970foundations},
which uses convex optimization techniques on different slices of the
tensor. However, a major disadvantage of ALS is that it does not perform
well in the presence of highly noisy measurements. Moreover, its theoretical
properties are not well understood. Here we derive and analyze an
approximate message passing (AMP) algorithm for optimal Bayesian recovery
of arbitrarily shaped, high-order low-rank tensors buried in noise.
As a result, we obtain an AMP algorithm that both out-performs ALS
and admits an analytic theory of its performance limits.

AMP algorithms have a long history dating back to early work on the
statistical physics of perceptron learning \cite{mezard1989space,kabashima2004bp}
(see \cite{advani2013statistical} for a review). The term AMP was
coined by Donoho, Maleki and Montanari in their work on compressed
sensing \cite{donoho2009message} (see also \cite{kabashima2009typical,rangan2009asymptotic,ganguli2010statistical,ganguli2010short,advani2016statistical,advani2016equivalence}
for replica approaches to compressed sensing and high dimensional
regression). AMP approximates belief propagation in graphical models
and a rigorous analysis of AMP was carried out in \cite{bayati2011dynamics}.
For a rank-one matrix estimation problem, AMP was first introduced
and analyzed in \cite{rangan2012iterative}. This framework has been
extended in a beautiful body of work by Krzakla and Zdeborova and
collaborators to various low-rank matrix factorization problems in
\cite{lesieur2015mmse,lesieur2017constrained,lesieur2017statistical,lesieur2015phase}).
Also, recently low-rank tensor decomposition through AMP was studied
in \cite{lesieur2017statistical}, but their analysis was limited
to symmetric tensors which are then necessarily cubic in shape.

However, tensors that occur naturally in the wild are almost never
cubic in shape, nor are they symmetric. The reason is that the $p$
different modes of an order $p$ tensor correspond to measurements
across very different modalities, resulting in very different numbers
of dimensions across modes, yielding highly irregularly shaped, non-cubic
tensors with no symmetry properties. For example in EEG studies $3$
different tensor modes could correspond to time, spatial scale, and
electrodes \cite{acar2007multiway}. In fMRI studies the modes could
span channels, time, and patients \cite{hunyadi2017tensor}. In neurophysiological
measurements they could span neurons, time, and conditions \cite{seely2016tensor}
or neurons, time, and trials \cite{williams2017unsupervised}. In
studies of visual cortex, modes could span neurons, time and stimuli
\cite{rabinowitz2015attention}.

Thus, given that tensors in the wild are almost never cubic, nor symmetric,
to bridge the gap between theory and experiment, we go beyond prior
work to derive and analyze Bayes optimal AMP algorithms for \textit{arbitrarily
shaped} high order and low rank tensor decomposition with \textit{different}
priors for different tensor modes, reflecting their different measurement
types. We find that the low-rank decomposition problem admits two
phase transitions separating three qualitatively different inference
regimes: (1) the easy regime at low noise where AMP works, (2) the
hard regime at intermediate noise where AMP fails but the ground truth
tensor is still possible to recover, if not in a computationally tractable
manner, and (3) the impossible regime at high noise where it is believed
no algorithm can recover the ground-truth low rank tensor.

From a theoretical perspective, our analysis reveals several surprises
relative to the analysis of symmetric cubic tensors in \cite{lesieur2017statistical}.
First, for symmetric tensors, it was shown that the easy inference
regime \textit{cannot} exist, \textit{unless} the prior over the low
rank factor has non-zero mean. In contrast, for non-symmetric tensors,
one tensor mode \textit{can} have zero mean \textit{without} destroying
the existence of the easy regime, as long as the other modes have
non-zero mean. Furthermore, we find that in the space of all possible
tensor shapes, the hard regime has the largest width along the noise
axis when the shape is cubic, thereby indicating that tensor shape
can have a strong effect on inference performance, and that cubic
tensors have highly non-generic properties in the space of all possible
tensor shapes.

Before continuing, we note some connections to the statistical mechanics
literature. Indeed, AMP is closely equivalent to the TAP equations
and the cavity method \cite{thouless1977solution,crisanti1995thouless}
in glassy spin systems. Furthermore, the posterior distribution of
noisy tensor factorization is equivalent to $p$-spin magnetic systems
\cite{crisanti1992sphericalp}, as we show below in section \ref{subsec:Bayesian-inference}.
For Bayes-optimal inference, the phase space of the problem is reduced
to the Nishimori line \cite{nishimori2001statistical}. This ensures
that the system does not exhibit replica-symmetry breaking. Working
in the Bayes-optimal setting thus significantly simplifies the statistical
analysis of the model. Furthermore, it allows theoretical insights
into the inference phase-transitions, as we shall see below. In practice,
for many applications the prior or underlying rank of the tensors
are not known \textit{a-priori}. The algorithms we present here can
also be applied in a non Bayes-optimal setting, where the parametric
from of the prior can not be determined. In that case, the theoretical
asymptotics we describe here may not hold. However, approximate Bayesian-optimal
settings can be recovered through parameter learning using expectation-maximization
algorithms \cite{dempster1977maximum}. We discuss these consequences
in section \ref{sec:AMP-vs-ALS}. Importantly,the connection to the
statistical physics of magnetic systems allows the adaptation of many
tools and intuitions developed extensively in the past few decades,
see e.g. \cite{zdeborova2016statistical}. We discuss more connections
to statistical mechanics as we proceed below.

\section{Low rank decomposition using approximate message passing}

In the following we define the low-rank tensor decomposition problem
and present a derivation of AMP algorithms designed to solve this
problem, as well as a dynamical mean field theory analysis of their
performance. A full account of the derivations can be found in the
supplementary material.

\subsection{Low-rank tensor decomposition\label{subsec:TCA}}

\begin{figure}
\includegraphics[width=1\textwidth]{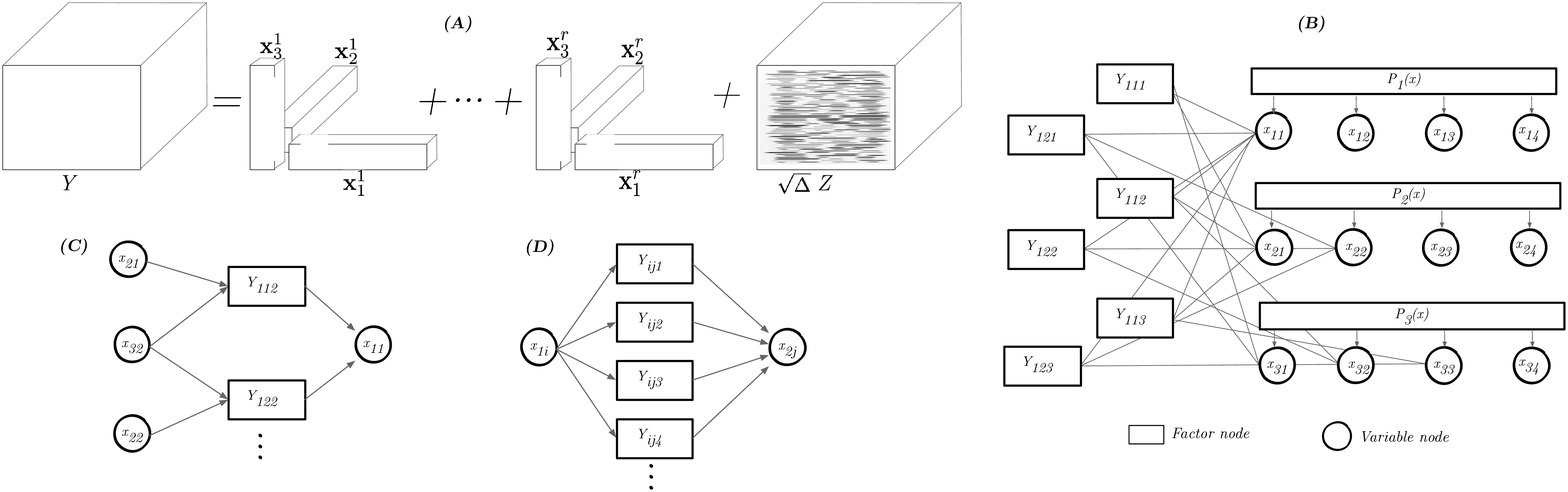}\caption{\emph{Low rank-decomposition of an order-$3$ spiked-tensor}. \textbf{(A)}
The observation tensor $Y$ is a sum of $r$ $rank-1$ tensors and
a noise tensor $\epsilon$ with variance $\Delta$. \textbf{(B)} Factor
graph for the decomposition of an order-3 tensor. \textbf{(C)} Incoming
messages into each variable node arrives from variable nodes connected
to the adjacent factor nodes. \textbf{(D)} Each node receives $N^{p-2}$
messages from each of the other variable nodes in the graph. \label{fig:decomposition}}
\end{figure}

Consider a general tensor $Y$ of order-$p$, whose components are
given by a set of $p$ indices, $Y_{i_{1},i_{2},...,i_{p}}$. Each
index $i_{\alpha}$ is associated with a specific \emph{mode} of the
tensor. The dimension of the mode $\alpha$ is $N_{\alpha}$ so the
index $i_{\alpha}$ ranges from $1,\ldots,N_{\alpha}$. If $N_{\alpha}=N$
for all $\alpha$ then the tensor is said to be \emph{cubic}. Otherwise
we define $N$ as the geometric mean of all dimensions $N=(\prod_{\alpha}^{p}N_{\alpha})^{1/p}$,
and denote $n_{\alpha}\equiv N_{\alpha}/N$ so that $\prod_{\alpha}^{p}n_{\alpha}=1$.
We employ the shorthand notation $Y_{i_{1},i_{2},...,i_{p}}\equiv Y_{a}$,
where $a=\{i_{1},\ldots,i_{p}\}$ is a set of $p$ numbers indicating
a specific element of $Y$. A rank-$1$ tensor of order-$p$ is the
outer product of $p$ vectors (order-$1$ tensors) $\prod_{1\leq\alpha\leq p}^{\otimes}\mathbf{x}_{\alpha},$where
$\mathbf{x_{\alpha}}\in\mathbb{R}^{N_{\alpha}}$. A rank-$r$ tensor
of order-$p$ has a special structure that allows it to be decomposed
into a sum of $r$ rank-$1$ tensors, each of order-$p$. The goal
of the rank decomposition is to find all $\mathbf{x}_{\alpha}^{\rho}\in\mathbb{R}^{N_{\alpha}}$
, for $\alpha=1,..p,$ and $\rho=1,\ldots,r$, given a tensor $Y$
of order-$p$ and rank-$r$. In the following, we will use $\mathbf{x}_{\alpha i}\in\mathbb{R}^{r}$
to denote the vector of values at each entry of the tensor, spanning
the $r$ rank-$1$ components. In a low-rank decomposition it is assumed
that $r<N$. In \emph{noisy} low-rank decomposition, individual elements
$Y_{a}$ are noisy measurements of a low-rank tensor {[}Figure \ref{fig:decomposition}.A{]}.
A comprehensive review on tensor decomposition can be found in \cite{kolda2009tensor}.

We state the problem of low-rank noisy tensor decomposition as follows:
Given a rank-$r$ tensor 
\begin{equation}
w_{a}=\frac{1}{N^{\frac{p-1}{2}}}\sum_{\rho=1}^{r}\prod_{\alpha}x_{\alpha i}^{\rho},\label{eq:w}
\end{equation}
we would like to find all the underlying factors $x_{\alpha i}^{\rho}$.
We note that we have used the shorthand notation $i=i_{\alpha}$ to
refer to the index $i_{\alpha}$ which ranges from $1$ to $N_{\alpha}$,
i.e. the dimensionality of mode $\alpha$ of the tensor.

Now consider a noisy measurement of the rank-$r$ tensor $w$ given
by 
\begin{equation}
Y=w+\sqrt{\Delta}\epsilon,\label{eq:eq:AWGN}
\end{equation}
where $\epsilon$ is a random noise tensor of the same shape as $w$
whose elements are distributed i.i.d according to a standard normal
distribution, yielding a total noise variance $\Delta\sim O(1)$ {[}Fig.
\ref{fig:decomposition}.A{]}. The underlying factors $x_{\alpha i}^{\rho}$
are sampled i.i.d from a \emph{prior} distribution $P_{\alpha}(x)$,
that may vary between the modes $\alpha$. This model is a generalization
of the spiked-tensor models studied in \cite{lesieur2017statistical,richard2014statistical}.

We study the problem in the thermodynamic limit where $N\to\infty$
while $r$,$n_{\alpha}\sim O(1)$. In that limit, the mean-field theory
we derive below becomes exact. The achievable performance in the decomposition
problem depends on the signal-to-noise ratio (SNR) between the underlying
low rank tensor (the signal) and the noise variance $\Delta$. In
eq. \eqref{eq:w} we have scaled the SNR (signal variance divided
by noise variance) with $N$ so that the SNR is proportional to the
ratio between the $O(N)$ unknowns and the $N^{p}$ measurements,
making the inference problem neither trivially easy nor always impossible.
From a statistical physics perspective, this same scaling ensures
that the posterior distribution over the factors given the data corresponds
to a Boltzmann distribution whose Hamiltonian has extensive energy
proportional to $N$, which is necessary for nontrivial phase transitions
to occur.

\subsection{Tensor decomposition as a Bayesian inference problem \label{subsec:Bayesian-inference}}

In Bayesian inference, one wants to compute properties of the posterior
distribution 
\begin{equation}
P(w\vert Y)=\frac{1}{Z(Y,w)}\prod_{\rho}^{r}\prod_{\alpha}^{p}\prod_{i}^{N}P_{\alpha}(x_{\alpha i}^{\rho})\prod_{a}P_{out}\cond{Y_{a}}{w_{a}}.\label{eq:post}
\end{equation}

Here $P_{out}\cond{Y_{a}}{w_{a}}$ is an element-wise output channel
that introduce independent noise into individual measurements. For
additive white Gaussian noise, the output channel in eq. \eqref{eq:post}
is given by 
\begin{equation}
\log P_{out}\cond{Y_{a}}{w_{a}}=g\cond{Y_{a}}{w_{a}}=\frac{1}{2\Delta}\left(Y_{a}-w_{a}\right)^{2}-\frac{1}{2}\log2\pi\Delta,\label{eq:cost}
\end{equation}
where $g(\cdot)$ is a quadratic \emph{cost function}. The denominator
$Z(Y,w)$, in \ref{eq:post} is a normalization factor, or the partition
function in statistical physics. In a Bayes-optimal setting, the priors
$P_{\alpha}(x)$, as well as the rank $r$ and the noise $\Delta$
are known.

The channel universality property \cite{lesieur2015mmse} states that
for low-rank decomposition problems, $r\ll N$, any output channel
is equivalent to simple additive white Gaussian noise, as defined
in eq. \eqref{eq:eq:AWGN}. Briefly, the output channel can be developed
as a power series in $w_{a}$. For low-rank estimation problems we
have $w_{a}\ll1$ {[}eq. \eqref{eq:w}{]}, and we can keep only the
leading terms in the expansion. One can show that the remaining terms
are equivalent to random Gaussian noise, with variance equal to the
inverse of the Fisher information of the channel {[}See supplementary
material for further details{]}. Thus, non-additive and non-Gaussian
measurement noise at the level of individual elements, can be replaced
with an effective additive Gaussian noise, making the theory developed
here much more generally applicable to diverse noise scenarios.

The motivation behind the analysis below, is the observation that
the posterior \eqref{eq:post}, with the quadratic cost function \eqref{eq:cost}
is equivalent to a Boltzmann distribution of a magnetic system at
equilibrium, where $\text{\ensuremath{\mathbf{x}_{\alpha i}\in\mathbb{R}^{r}}}$
can be though of as the $r$-dimensional vectors of a spherical (xy)-spin
model \cite{crisanti1992sphericalp}.

\subsection{Approximate message passing on factor graphs}

To solve the problem of low-rank decomposition we frame the problem
as a graphical model with an underlying bipartite factor graph. The
variable nodes in the graph represent the $rN\sum_{\alpha}n_{\alpha}$
unknowns $x_{\alpha i}^{\rho}$ and the $N^{p}$ factor nodes correspond
to the measurements $Y_{a}$. The edges in the graph are between factor
node $Y_{a}$ and the variable nodes in the neighbourhood $\partial a$
{[}Figure \ref{fig:decomposition}.B{]}. More precisely, for each
factor node $a=\{i_{1},i_{2},...,i_{p}\}$, the set of variable nodes
in the neighbourhood $\partial a$ are precisely $\{\boldsymbol{x}_{1i_{1}},\boldsymbol{x}_{2i_{2}},...,\boldsymbol{x}_{pi_{p}}\}$,
where each $\boldsymbol{x}_{\alpha i_{\alpha}}\in\mathbb{R}^{r}$.
Again, in the following we will use the shorthand notation $\boldsymbol{x}_{\alpha i}$
for $\boldsymbol{x}_{\alpha i_{\alpha}}$. The state of a variable
node is defined as the marginal probability distribution $\eta_{\alpha i}(\mathbf{x})$
for each of the $r$ components of the vectors $\mathbf{x}_{\alpha i}\in\mathbb{R}^{r}$
. The estimators $\hat{\mathbf{x}}_{\alpha i}\in\mathbb{R}^{r}$ for
the values of the factors $\mathbf{x}_{\alpha i}$ are given by the
means of each of the marginal distributions $\eta_{\alpha i}(\mathbf{x})$.

In the approximate message passing framework, the state of each node
(also known as a 'belief'), $\eta_{\alpha i}(x)$ is transmitted to
all other variable nodes via its adjacent factor nodes {[}Fig. \ref{fig:decomposition}.C{]}.
The state of each node is then updated by marginalizing over all the
incoming messages, weighted by the cost function and observations
in the factor nodes they passed on the way in: 
\begin{equation}
\eta_{\alpha i}(\mathbf{x})=\frac{P_{\alpha}(\mathbf{x})}{Z_{\alpha i}}\prod_{a\in\partial\alpha i}\prod_{\beta j\in\partial a\setminus\alpha i}Tr_{x_{\beta j}}\eta_{\beta j}(x_{\beta j})e^{g(y_{a},w_{a})}.\label{eq:eta}
\end{equation}
Here $P_{\alpha}(\mathbf{x})$ is the prior for each factor $\mathbf{x}_{\alpha i}$
associated with mode $\alpha$, and $Z_{\alpha i}=\intop d\mathbf{x}\eta_{\alpha i}(\mathbf{x})$
is the partition function for normalization. The first product in
\eqref{eq:eta} spans all factor nodes adjacent to variable node $\alpha i$.
The second product is over all variable nodes adjacent to each of
the factor nodes, excluding the target node $\alpha i$. The trace
$Tr_{x_{\beta j}}$ denotes the marginalization of the cost function
$g(y_{a},w_{a})$ over all incoming distributions.

The mean of the marginalized posterior at node $\alpha i$ is given
by

\begin{equation}
\hat{\mathbf{x}}_{\alpha i}=\int dx\eta_{\alpha i}(\mathbf{x})\mathbf{x}\in\mathbb{R}^{r},\label{eq:xhat}
\end{equation}
and its covariance is 
\begin{equation}
\hat{\sigma}_{\alpha i}^{2}=\int dx\eta_{\alpha i}(x)xx^{T}-\hat{x}_{\alpha i}\hat{x}_{\alpha i}^{T}\in\mathbb{R}^{r\times r}.\label{eq:sighat}
\end{equation}

Eq. \eqref{eq:eta} defines an iterative process for updating the
beliefs in the network. In what follows, we use mean-field arguments
to derive iterative equations for the means and covariances of the
these beliefs in \eqref{eq:xhat}-\eqref{eq:sighat}. This is possible
given the assumption that incoming messages into each node are probabilistically
independent. Independence is a good assumption when short loops in
the underlying graphical model can be neglected. One way this can
occur is if the factor graph is sparse \cite{yedidia2001bethe,mezard2009information}.
Such graphs can be approximated by a directed acyclic graph; in statistical
physics this is known as the Bethe approximation \cite{bethe1935statistical}.
Alternatively, in low-rank tensor decomposition, the statistical independence
of incoming messages originates from weak pairwise interactions that
scale as $w\sim N^{-(p-1)/2}$. Loops correspond to higher order terms
interaction terms, which become negligible in the thermodynamic limit
\cite{bayati2011dynamics,zdeborova2016statistical}.

Exploiting these weak interactions we construct an accurate mean-field
theory for AMP. Each node $\alpha i$ receives $N^{p-2}$ messages
from every node $\beta j$ with $\beta\neq\alpha$, through all the
factor nodes that are connected to both nodes,$\{y_{b}\vert b\in\partial\alpha i\cup\partial\beta j\}$
{[}Fig. \ref{fig:decomposition}.D{]}. Under the independence assumption
of incoming messages, we can use the central limit theorem to express
the state of node $\alpha j$ in \eqref{eq:eta} as 
\begin{equation}
\eta_{\alpha i}(\mathbf{x})=\frac{P_{\alpha}(\mathbf{x})}{Z_{\alpha}(A_{\alpha i},\mathbf{u}_{\alpha i})}\prod_{\beta j\neq\alpha i}\exp\left(-\mathbf{x}^{T}A_{\beta j}\mathbf{x}+\mathbf{u}_{\beta j}^{T}\mathbf{x}\right),\label{eq:eta_MF}
\end{equation}
where $A_{\beta}^{-1}\mathbf{u}_{\beta j}$ and $A_{\beta}^{-1}$
are the mean and covariance of the local incoming messages respectively.
The distribution is normalized by the partition function 
\begin{equation}
Z_{\alpha}(A,\mathbf{u})=\intop dxP_{\alpha}(x)\exp\left[\left(\mathbf{u}^{T}\mathbf{x}-\mathbf{x}^{T}A\mathbf{x}\right)\right].
\end{equation}
The mean and covariance of the distribution, eq. \eqref{eq:xhat}
and \eqref{eq:sighat} are the moments of the partition function

\begin{equation}
\hat{\mathbf{x}}_{\alpha i}=\frac{\partial}{\partial\mathbf{u}_{\alpha i}}\log Z_{\alpha},\;\hat{\sigma}_{\alpha i}=\frac{\partial^{2}}{\partial\mathbf{u}_{\alpha i}\partial\mathbf{u}_{\alpha i}^{T}}\log Z_{\alpha}.\label{eq:x_and_s}
\end{equation}
Finally, by expanding $g(Y_{a},w_{a}$) in eq. \eqref{eq:eta} to
quadratic order in $w$, and averaging over the posterior, one can
find a self consistent equation for $A_{\alpha i}$ and $\mathbf{u}_{\alpha i}$
in terms of $\mathbf{x}_{\alpha i}$ and $Y$ {[}see supplemental
material for details{]}.

\subsection{AMP algorithms}

Using equations \eqref{eq:x_and_s}, and the self-consistent equations
for $A_{\alpha i}$ and $\mathbf{u}_{\alpha i}$, we construct an
iterative algorithm whose dynamics converges to the solution of the
self-consistent equations {[} see supplemental material for details{]}.
The resulting update equations for the parameters are 
\begin{align}
\mathbf{u}_{\alpha i}^{t} & =\frac{n_{\alpha}}{\Delta N^{(p-1)/2}}\sum_{a\in\partial\alpha i}Y_{a}\left(\prod_{(\beta,j)\in\partial b\setminus(\alpha,i)}^{\odot}\hat{\mathbf{x}}_{\beta j}^{t}\right)-\frac{1}{\Delta}\hat{\mathbf{x}}_{\alpha i}^{t-1}\sum_{\beta\neq\alpha}\Sigma_{\beta}^{t}\odot D_{\alpha\beta}^{t,t-1}\label{eq:alg_u}\\
A_{\alpha}^{t} & =\frac{1}{\Delta}\prod_{\beta\neq\alpha}^{\odot}\left(\frac{1}{n_{\beta}N}\sum_{j=1}^{N}\hat{\mathbf{x}}_{\beta j}^{t}\hat{\mathbf{x}}_{\beta j}^{tT}\right)\label{eq:alg_A}\\
\mathbf{\hat{x}}_{\alpha i}^{t+1} & =\frac{\partial}{\partial\mathbf{u}_{\alpha i}^{t}}\log Z_{\alpha}(A_{\alpha}^{t},\mathbf{u}_{\alpha i}^{t})\label{eq:alg_x}\\
\hat{\sigma}_{\alpha i}^{t+1} & =\frac{\partial^{2}}{\partial\mathbf{u}_{\alpha i}^{t}\partial\mathbf{u}_{\alpha i}^{tT}}\log Z_{\alpha}(A_{\alpha}^{t},\mathbf{u}_{\alpha i}^{t}),\label{eq:alg_s}
\end{align}
The second term on the RHS of \eqref{eq:alg_u}, is given by 
\begin{equation}
D_{\alpha\beta}^{t,t-1}=\prod_{\gamma\neq\alpha,\beta}^{\odot}\left(\frac{1}{N}\sum_{k}\hat{\mathbf{x}}_{\gamma k}^{t}\hat{\mathbf{x}}_{\gamma k}^{t-1,T}\right),\;\;\Sigma_{\alpha}^{t}=N^{-1}\sum_{i}\hat{\sigma}_{\alpha_{i}}^{t}.
\end{equation}
This term originates from the the exclusion the target node $\alpha i$
from the product in equations \eqref{eq:eta} and \eqref{eq:eta_MF}.
In statistical physics it corresponds to an Onsager reaction term
due to the removal of the node yielding a cavity field \cite{mezard1985j}.
In the above, the notations $\odot$, $\prod^{\odot}$ denote component-wise
multiplication between two, and multiple tensors respectively.

Note that in the derivation of the iterative update equation above,
we have implicitly used the assumption that we are in the Bayes-optimal
regime which simplifies eq. \eqref{eq:alg_u}-\eqref{eq:alg_s} {[}see
supplementary material for details{]}. The AMP algorithms can be derived
without the assumption of Bayes-optimality, resulting in a slightly
more complicated set of algorithms {[}See supplementary material for
details{]}. However, further analytic analysis, which is the focus
of this current work, and the derivation of the dynamic mean-field
theory which we present below is applicable in the Bayes-optimal regime,
were there is no replica-symmetry breaking, and the estimators are
self-averaging. Once the update equations converge, the estimates
for the factors $\mathbf{x}_{\alpha i}$ and their covariances are
given by the fixed point value of equations \eqref{eq:alg_x} and
\eqref{eq:alg_s} respectively. A statistical treatment for the convergence
in typical settings is presented in the following section.

\subsection{Dynamic mean-field theory }

To study the performance of the algorithm defined by eq. \eqref{eq:alg_u}-\eqref{eq:alg_s},
we use another mean-field approximation that estimates the evolution
of the inference error. As before, the mean-field becomes exact in
the thermodynamic limit. We begin by defining order parameters that
measure the correlation of the estimators $\hat{\mathbf{x}}_{\alpha i}^{t}$
with the ground truth values $\mathbf{x}_{\alpha i}$ for each mode
$\alpha$ of the tensor 
\begin{equation}
M_{\alpha}^{t}=\left(n_{\alpha}N\right)^{-1}\sum_{i=1}^{N_{\alpha}}\hat{\mathbf{x}}_{\alpha i}^{t}\mathbf{x}_{\alpha i}^{T}\in\mathbb{R}^{r\times r}.\label{eq:M}
\end{equation}
Technically, the algorithm is permutation invariant, so one should
not expect the high correlation values to necessarily appear on the
diagonal of $M_{\alpha}^{t}$. In the following, we derive an update
equation for $M_{\alpha}^{t}$, which will describe the performance
of the algorithm across iterations.

An important property of Bayes-optimal inference is that there is
no statistical difference between functions operating on the ground
truth values, or on values sampled uniformly from the posterior distribution.
In statistical physics this property is known as one of the \emph{Nishimori
conditions} \cite{nishimori2001statistical}. These conditions allow
us to derive a simple equation for the update of the order parameter
\eqref{eq:M}. For example, from \eqref{eq:alg_A} one easily finds
that in Bayes-optimal settings $A_{\alpha}^{t}=\bar{M}_{\alpha}^{t}.$
Furthermore, averaging the expression for $\mathbf{u}_{\alpha i}$
over the posterior, we find that {[} supplemental material{]}

\begin{equation}
\mathbb{E}_{P(W\vert Y)}\left[\mathbf{u}_{\alpha i}^{t}\right]=\bar{M}_{\alpha}^{t}x_{\alpha i},\label{eq:mean_u}
\end{equation}
where 
\begin{equation}
\bar{M}_{\alpha}^{t}\equiv\frac{n_{\alpha}}{\Delta}\prod_{\beta\neq\alpha}^{\odot}M_{\beta}^{t}.\label{eq:Mbar}
\end{equation}
Similarly, the covariance matrix of $\mathbf{u}_{\alpha i}$ under
the posterior is 
\begin{equation}
COV_{P(W\vert Y)}\left[\mathbf{u}_{i\alpha}^{t}\right]=\bar{M}_{\alpha}^{t}.\label{eq:cov_u}
\end{equation}

Finally, using eq. \eqref{eq:alg_x} for the estimation of $\hat{\mathbf{x}}_{\alpha i}$,
and the definition of $M_{\alpha}^{t}$ in \eqref{eq:M} we find a
dynamical equation for the evolution of the order parameters $M_{\alpha}^{t}$:
\begin{equation}
M_{\alpha}^{t+1}=\mathbb{E}_{P_{\alpha}(\mathbf{x}),z}\left[f_{\alpha}\left(\bar{M}_{\alpha}^{t},\bar{M}_{\alpha}^{t}x_{\alpha i}+\sqrt{\bar{M}_{\alpha}^{t}}\mathbf{z}\right)x_{\alpha i}^{T}\right],\label{eq:DMFT_SE}
\end{equation}
where $f_{\alpha}\equiv\frac{\partial}{\partial\mathbf{u}}\log Z_{\alpha}(A,\mathbf{u})$
is the estimation of $\mathbf{\hat{x}}_{\alpha i}^{t+1}$ from \eqref{eq:alg_x}.
The average in \eqref{eq:DMFT_SE} is over the prior $P_{\alpha}(\mathbf{x})$
and over the standard Gaussian variables $\mathbf{z}\in\mathbb{R}^{r}$.
The average over $\mathbf{z}$ represents fluctuations in the mean
$\bar{M}_{\alpha}^{t}x_{\alpha i}$ in \eqref{eq:mean_u}, due to
the covariance $\bar{M}_{\alpha}^{t}$ in \eqref{eq:cov_u}.

Finally, the performance of the algorithm is given by the fixed point
of \eqref{eq:DMFT_SE}, 
\begin{equation}
M_{\alpha}^{*}=\mathbb{E}_{P_{\alpha}(\mathbf{x}),z}\left[f_{\alpha}\left(\bar{M}_{\alpha}^{*},\bar{M}_{\alpha}^{*}x_{\alpha i}+\sqrt{\bar{M}_{\alpha}^{*}}\mathbf{z}\right)x_{\alpha i}^{T}\right],\;\text{where }\bar{M}_{\alpha}^{*}\equiv\frac{n_{\alpha}}{\Delta}\prod_{\beta\neq\alpha}^{\odot}M_{\beta}^{*}.\label{eq:DMFT_FP}
\end{equation}

As we will see below, the inference error can be calculated from the
fixed point order parameters $M_{\alpha}^{*}$ in a straightforward
manner.

\section{Phase transitions in generic low-rank tensor decomposition }

The dynamics of $M_{\alpha}^{t}$ depend on the SNR via the noise
level $\Delta$. To study this dependence, we solve equations \eqref{eq:DMFT_SE}
and \eqref{eq:DMFT_FP} with specific priors. Below we present the
solution of using Gaussian priors. In the supplementary material we
also solve for Bernoulli and Gauss-Bernoulli distributions, and discuss
mixed cases where each mode of the tensor is sampled from a different
prior. Given our choice of scaling in \eqref{eq:w}, we expect phase
transitions at $O(1)$ values of $\Delta$, separating three regimes
where inference is: (1) easy at small $\Delta$; (2) hard at intermediate
$\Delta$; and (3) impossible at large $\Delta$. For simplicity we
focus on the case of rank $r=1$, where the order parameters $M_{\alpha}^{t}$
in \eqref{eq:M} become scalars, which we denote $m_{\alpha}^{t}$.

\subsection{Solution with Gaussian priors}

We study the case where $x_{\alpha i}$ are sampled from normal distributions
with mode-dependent mean and variance $P_{\alpha}(x)\sim\mathcal{N}(\mu_{\alpha},\sigma_{\alpha}^{2}).$
The mean-field update equation \eqref{eq:DMFT_SE} can be written
as 
\begin{equation}
m_{\alpha}^{t+1}=\frac{\frac{\mu_{\alpha}^{2}}{\sigma_{\alpha}^{2}}+\left(\sigma^{2}+\mu^{2}\right)\bar{m}_{\alpha}^{t}}{\sigma^{-2}+\bar{m}_{\alpha}^{t}},\label{eq:DMFT_gauss}
\end{equation}
where $\bar{m}_{\alpha}^{t}\equiv\Delta^{-1}n_{\alpha}\prod_{\beta\neq\alpha}m_{\beta}$
, as in \eqref{eq:Mbar}. We define the average inference error for
all modes 
\begin{equation}
MSE=\frac{1}{p}\sum_{\alpha}\frac{\left|\hat{x}_{\alpha}-x_{\alpha}\right|^{2}}{2\sigma_{\alpha}^{2}}=\frac{1}{p}\sum_{\alpha}\left(1+\frac{\mu_{\alpha}^{2}}{\sigma_{\alpha}^{2}}-\frac{1}{\sigma_{\alpha}^{2}}m_{\alpha}^{*}\right),\label{eq:mse}
\end{equation}

where $m_{\alpha}^{*}$ is the fixed point of eq. \eqref{eq:DMFT_gauss}.
Though we focus here on the $r=1$ case for simplicity, the theory
is equally applicable to higher-rank tensors.

\begin{figure}
\includegraphics{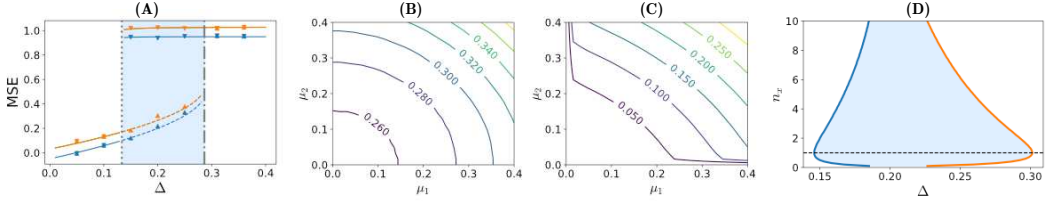}\vspace{-1em}
\caption{\label{fig:PT}\emph{Phase transitions in the inference of arbitrary
order-3 rank-$1$ tensors}. \textbf{(A)} MSE at the fixed point plotted
against the noise level $\Delta$. Shaded blue area marks the bi-stability
regime $\Delta_{alg}<\Delta<\Delta_{dyn}$. Solid (dashed) lines show
theoretically predicted MSE for random (informed) initializations.
Points are obtained from numerical simulations {[}$\sigma_{\alpha}=1$,
$\mu_{1}=\mu_{2}=0.1$ (blue), $\mu_{3}=0.3$ (orange), $N=500$,
$n_{\alpha}=1${]}. \textbf{(B) }Contours of $\Delta_{dyn}$ as a
function of the two non-zero means of modes $\text{\ensuremath{\alpha=1,2}}$
{[}$\mu_{3}=0$, $\sigma_{\alpha}=1]$. As either of the two nonzero
means increases, $\Delta_{dyn}$ increases with them, reflecting an
increase in the regime of noise level $\Delta$ over which inference
is easy. Importantly, the transition $\Delta_{dyn}$ is finite even
when only one prior has non zero mean. \textbf{(C)} Same as (B) but
for $\Delta_{alg}$. Again, as either mean increases, $\Delta_{alg}$
increases also, reflecting a delay in the onset of the impossible
regime as the noise level $\Delta$ increases. The algorithmic phase
transition is finite when at most one prior has zero mean. \textbf{(D)}
Lower and higher transition points $\Delta_{Alg}$ (blue) and $\Delta_{Dyn}$
(orange) as a function of tensor \textit{shape}. The ratios between
the mode dimensions are $n_{\alpha}=\{1,n_{x},1/n_{x}\}$. The width
of the bi-stable or hard inference regime is widest at the cubic point
where $n_{x}=1$. }
\vspace{-1em}
 
\end{figure}

Solutions to the theory in \eqref{eq:DMFT_gauss} and \eqref{eq:mse}
are plotted in Fig. \ref{fig:PT}.A together with numerical simulations
of the algorithm \eqref{eq:alg_u}-\eqref{eq:alg_s} for order-3 tensors
generated randomly according to \eqref{eq:eq:AWGN}. The theory and
simulations match perfectly. The AMP dynamics for general tensor decomposition
is qualitatively similar to that of rank-$1$ symmetric matrix and
tensor decompositions, despite the fact that such symmetric objects
possess only one mode. As a consequence, the space of order parameters
for these two problems is only one-dimensional; in contrast for the
general case we consider here, it is $p$-dimensional. Indeed, the
$p=3$ order parameters are all simultaneously and correctly predicted
by our theory.

For low levels of noise, the iterative dynamics converge to a stable
fixed point of \eqref{eq:DMFT_gauss} with low MSE. As the noise increases
beyond a bifurcation point $\Delta_{alg}$, a second stable fixed
point emerges with $m_{\alpha}^{*}\ll1$ and $MSE\approx$1. Above
this point AMP may not converge to the true factors. The basin of
attraction of the two stable fixed points are separated by a $p-1$
dimensional sub-manifold in the $p$-dimensional order parameter space
of $m_{\alpha}$. If the initial values $x_{\alpha i}^{0}$ have sufficiently
high overlap with the true factors $x_{\alpha i}$, then the AMP dynamics
will converge to the low error fixed point; we refer to this as the
informative initialization, as it requires prior knowledge about the
true structure. For uninformative initializations, the dynamics will
converge to the high error fixed point almost surely in the thermodynamic
limit.

At a higher level of noise, $\Delta_{dyn}$, another pitchfork bifurcation
occurs and the high error fixed point becomes the \emph{only} stable
point. With noise levels $\Delta$ above $\Delta_{dyn}$, the dynamic
mean field equations will always converge to a high error fixed point.
In this regime AMP cannot overcome the high noise and inference is
impossible.

From eq. \eqref{eq:DMFT_gauss}, it can be easily checked that if
the prior means $\mu_{\alpha}=0$, $\forall\alpha$ then the high
error fixed point with $m_{\alpha}=0$ is \emph{stable} for any finite
$\Delta$. This implies that $\Delta_{alg}=0$, and there is no easy
regime for inference, so $AMP$ with uninformed initialization will
never find the true solution. This difficulty was previously noted
for the low-rank decomposition of symmetric tensors \cite{lesieur2017statistical},
and it was further shown there that the prior \emph{must} be non-zero
for the existence of an easy inference regime. However, for general
tensors there is higher flexibility; one mode $\alpha$ \emph{can}
have a zero mean without destroying the existence of an easy regime.
To show this we solved \eqref{eq:DMFT_gauss}, with different prior
means for different modes and we plot the phase boundaries in Fig.
\ref{fig:PT}.B-C. For the $p=3$ case, $\Delta_{dyn}$ is finite
even if two of the priors have zero mean. Interestingly, the algorithmic
transition $\Delta_{alg}$ is finite if at most one prior is has zero
mean. Thus, the general tensor decomposition case is qualitatively
different than the symmetric case in that an easy regime can exist
even when a tensor mode has zero mean.

\subsection{Non-cubic tensors}

The shape of the tensor, defined by the different mode dimensions
$n_{\alpha}$, has an interesting effect on the phase transition boundaries,
which can be studied using \eqref{eq:DMFT_gauss}. In figure \ref{fig:PT}.D
the two transitions, $\Delta_{alg}$ and $\Delta_{dyn}$ are plotted
as a function of the shape of the tensor. Over the space of all possible
tensor shapes, the boundary between the hard and impossible regimes,
$\Delta_{dyn}$ is maximized, or pushed furthest to the right in Fig.
\ref{fig:PT}.D, when the shape takes the special cubic form where
all dimensions are equal $n_{\alpha}=1,\;\forall\alpha$. This diminished
size of the impossible regime at the cubic point can be understood
by noting the cubic tensor has the highest ratio between the number
of observed data points $N^{p}$ and the number of unknowns $rN\sum_{\alpha}n_{\alpha}$.

Interestingly the algorithmic transition is lowest at this point.
This means that although the ratio of observations to unknowns is
the highest, algorithms may not converge, as the width of the hard
regime is maximized. To explain this observation, we note that in
\eqref{eq:Mbar}, the noise can be rescaled independently in each
mode by defining $\Delta\to\Delta_{\alpha}=\Delta/n_{\alpha}$. It
follows that for non-cubic tensors the worst case effective noise
across modes will be necessarily higher than in the cubic case. As
a consequence, moving from cubic to non-cubic tensors lowers the minimum
noise level $\Delta_{alg}$ at which the uninformative solution is
stable, thereby extending the hard regime to the left in Fig. \ref{fig:PT}.D.

\vspace{-0.5em}

\section{Bayesian AMP compared to maximum a-posteriori (MAP) methods \label{sec:AMP-vs-ALS}}

\vspace{-0.5em}
 We now compare the performance of AMP to one of the most commonly
used algorithms in practice, namely alternating least squares (ALS)
\cite{kolda2009tensor}. ALS is motivated by the observation that
optimizing one mode while holding the rest fixed is a simple least-squares
subproblem \cite{harshman1970foundations,carroll1970analysis}. Typically,
ALS performs well at low noise levels, but here we explore how well
it compares to AMP at high noise levels, in the scaling regime defined
by defined by \eqref{eq:w} and \eqref{eq:eq:AWGN}, where inference
can be non-trivial.

\begin{figure}
\center\includegraphics[width=0.9\textwidth]{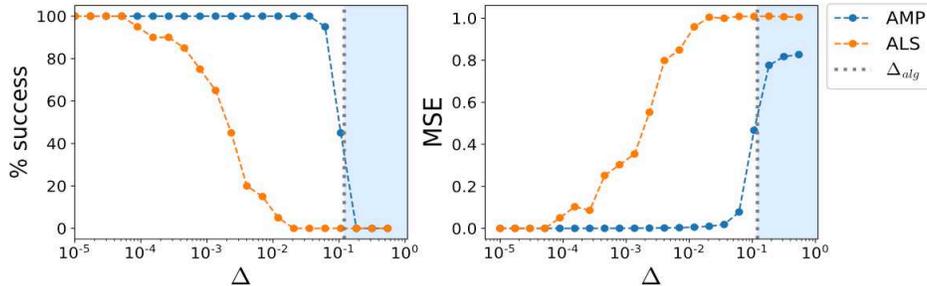}\caption{\emph{Comparing AMP and ALS }. \textbf{Left}: The percentage out of
50 simulations that converged to the low error solution as a function
of the noise $\Delta$ \textbf{Right}: MSE averaged over the 50 simulations.
In both figures, the vertical dashed line is the theoretical predication
of the algorithmic transition point $\Delta_{alg}$. {[}$p=3$, $r=1$,
$\sigma_{\alpha}=1$, $\mu_{\alpha}=0.2$, $n_{\alpha}=\{1,\frac{8}{10},\frac{10}{8}\}$,
$N=500$ {]} \label{fig:AMP-vs-ALS}}
\vspace{-1em}
 
\end{figure}

In Fig. \ref{fig:AMP-vs-ALS} we compare the performance of ALS with
that of AMP on the same underlying large ($N=500$) tensors with varying
amounts of noise. First, we note that that ALS does not exhibit a
sharp phase transition, but rather a smooth cross-over between solvable
and unsolvable regimes. Second, the robustness of ALS to noise is
much lower than that of AMP. This difference is more substantial as
the size of the tensors, $N$, is increased {[}data not shown{]}.

One can understand the difference in performance by noting that ALS
is like a MAP estimator, while Bayesian AMP attempts to find the minimal
mean square error (MMSE) solution. AMP does so by marginalizing probabilities
at every node. Thus AMP is expected to produce better inferences when
the posterior distribution is rough and dominated by noise. From a
statistical physics perspective, ALS is a zero-temperature method,
and so it is subject to replica symmetry breaking. AMP on the other
hand is Bayes-optimal and thus operates at the Nishimori temperature
\cite{nishimori2001statistical}. At this temperature the system does
not exhibit replica symmetry breaking, and the true global ground
state can be found in the easy regime, when $\Delta<\Delta_{alg}$.

\vspace{-0.05em}

\section{Summary}

\vspace{-0.5em}
 In summary, our work partially bridges the gap between theory and
practice by creating new AMP algorithms that can flexibly assign different
priors to different modes of a high-order tensor, thereby enabling
AMP to handle arbitrarily shaped high order tensors that actually
occur in the wild. Moreover, our theoretical analysis reveals interesting
new phenomena governing how irregular tensor shapes can strongly affect
inference performance and the positions of phase boundaries, and highlights
the special, non-generic properties of cubic tensors. Finally, we
hope the superior performance of our flexible AMP algorithms relative
to ALS will promote the adoption of AMP in the wild. Code to reproduce
all simulations presented in this paper is available at \emph{https://github.com/ganguli-lab/tensorAMP.}

\vspace{-0.5em}

\section*{Acknowledgments}

We thank Alex Williams for useful discussions. We thank the Center
for Theory of Deep Learning at the Hebrew University (J.K), and the
Burroughs-Wellcome, McKnight, James S.McDonnell, and Simons Foundations,
and the Office of Naval Research and the National Institutes of Health
(S.G) for support. \vspace{-0.5em}

\appendix

\global\long\def\Z{\mathcal{Z}}%
\global\long\def\H{\mathcal{H}}%
\global\long\def\O{O}%

\global\long\def\E{\mathbb{E}}%
\global\long\def\R{\mathcal{\mathbb{R}}}%
\global\long\def\cond#1#2{\left(#1\left|#2\right.\right)}%

\section*{Appendix}

In the following sections, we derive the approximate message passing
(AMP) algorithms for arbitrary low-rank tensors. The derivation follows
similar lines as the derivation in \cite{lesieur2017constrained,zdeborova2016statistical}
for the matrix $p=2$ case, only here the model is generalized for
higher modes p\textgreater 2. We then derive dynamical mean field
theory (also known as state evolution) and find the phase transition
of the inference problem using non-linear analysis on the recursive
dynamical equations of the order parameters. Lastly, we explicitly
solve the equations for several simple examples of mode-3 tensors
with a mixture of different prior distributions. For notational simplicity,
we start the derivation assuming all modes are of the same dimension,
$N$, which we assume to be in the thermodynamic limit, $N\to\infty.$
Then, we will generalize for noncubic tensors. We emphasize that even
in the cubic case the tensor is non-symmetric, and each mode is independent
and is iid drawn from a prior distribution, which is potentially different
for each mode.

\section{Message passing on factorized graph}

\subsection{Factor graph for tensor decomposition -- notations}

We consider a given low-rank tensor 
\begin{equation}
w_{a}^{0}=\frac{1}{N^{\frac{p-1}{2}}}\sum_{\rho=1}^{r}\prod_{\alpha}x_{\alpha i}^{0\rho}.
\end{equation}

Here the vectors $\mathbf{x}_{\alpha i}^{0}\in\R^{r}$ denote the
\emph{ground-truth} values to the estimation problem. Each entry of
the tensor is denoted with a lower-case latin letter $\{a,b,c,...\}$.
The notation stands for the set of $p$ indices that define that tensor
element,
\begin{equation}
a=\{i_{1},i_{2},...i_{p}\}.
\end{equation}

However, we only have access to noisy measurements of the ground-truth
vectors, denoted by 
\begin{equation}
Y_{a}=w_{a}^{0}+\sqrt{\Delta}\epsilon_{a},
\end{equation}
where $\epsilon_{a}$ is a random tesnsor, whose elements are i.i.d.
gaussians with zero mean and unit variance. We assume no covariance
between two measurements, $\mathbb{E}(\epsilon_{a}\epsilon_{b})=0$
$\forall a\neq b$.

The goal of the low rank decomposition is to find the estimators $\mathbf{\hat{x}}_{\alpha i}\in\R^{r}$
that minimize the mean square error
\[
\hat{x}=\arg\min_{x}\sum_{\alpha}\sum_{i}\left(x_{\alpha i}-x_{\alpha i}^{0}\right)^{2}.
\]
To solve the Bayesian inference problem, using message passing, we
frame it as a bipartite graphical model. Each of the variable nodes
corresponds to an estimator $\mathbf{x}_{\alpha i}$ {[}See figure
1.b in the main text{]}. We use the notation $\partial a$ to denote
all the neighboring nodes to $a$ and the notation $\partial a\backslash i\alpha$
to denote all the neighboring variable nodes adjacent to $a$, excluding
the node $\alpha i$. The cardinality of the set of all factor points
is $\left|\{a\}\right|=N^{P}$.

Each variable point on the graph is connected to $N^{p-1}$ factor
nodes. The set of neighboring factor nodes that is connected to the
variable node $\alpha i$ is denoted as
\begin{equation}
\partial\alpha i=\left\{ a\vert\alpha i\in a\right\} .
\end{equation}

\subsection{Weakly connected graph}

An underlying assumption in belief propagation and message passing
algorithms is that the incoming messages into each node are statistically
independent. It can be achieved, for example in sparse graphs, where
at each node the graph can be approximately considered as a tree (directed
acyclic graph), without recurring loops. In the physics literature
such approximation is often referred to as Bethe Lattice. In the current
model this is possible due to the scaling of individual elements,
$w\sim N^{(1-p)/2}$, as defined in Eq. (1) in the main text. Since
correlations in the messages are due to loops in the underlying graph,
that pass through several nodes, we neglect them when the interactions
are sufficiently weak \cite{mezard1986sk}.

\subsection{Message passing}

We start by defining two different types of messages (beliefs): one
for messages outgoing from a variable node into a factor node $\eta$.
Messages going from factor nodes into variable nodes are denoted by
$\tilde{\eta}$. Messages are the marginal probabilities at each node,
measuring the posterior probability density of the estimator at that
source node. A message outgoing from the variable node $\alpha i$
to a factor node $a$ can be written in terms of the product of messages
originating from all its connected nodes excluding $a$,
\begin{equation}
\eta_{\alpha i\to a}(\mathbf{x}_{\alpha i})=\frac{P_{\alpha}(\mathbf{x}_{\alpha i})}{\Z_{\alpha i\to a}}\prod_{b\in\partial\alpha i\setminus a}\tilde{\eta}_{b\to\alpha i}(\mathbf{x}_{\alpha i}).\label{eq:outgoing_msg}
\end{equation}
The denominator $\Z_{\alpha i\to a}$ is a normalization factor 
\[
\Z_{\alpha i\to a}=Tr_{x_{\alpha i}}P_{\alpha}(\mathbf{x}_{\alpha i})\prod_{b\in\partial\alpha i\setminus a}\tilde{\eta}_{b\to\alpha i}(\mathbf{x}_{\alpha i}).
\]
Incoming messages into variable nodes are obtained by marginalizing
over distribution over all the messages. A message outgoing from a
factor node into a variable node is given by 
\begin{equation}
\tilde{\eta}_{b\to\alpha i}(\mathbf{x}_{\alpha i})=\frac{1}{\Z_{b\to\alpha i}}\prod_{\beta j\in\partial b\backslash\alpha}Tr_{x_{\beta j}}\eta_{\beta j\to b}(\mathbf{x}_{\beta j})\exp g\left(Y_{b},N^{\frac{1-p}{2}}w_{b}\right),
\end{equation}
where 
\begin{equation}
w_{b}\equiv\sum_{\rho=1}^{r}\prod_{\alpha}x_{\alpha i}^{\rho}.\label{eq:wb}
\end{equation}
The normalization factor in the denominator of eq. \eqref{eq:wb}
is given by
\begin{equation}
\Z_{b\to\alpha i}=Tr_{x_{\alpha i}}\prod_{\beta j\in\partial b\backslash\alpha}Tr_{x_{\beta j}}\eta_{\beta j\to b}(\mathbf{x}_{\beta j})\exp g\left(Y_{b},N^{\frac{1-p}{2}}w_{b}\right).
\end{equation}
The cost function $g(\cdot)$ at the exponent can be expanded as a
power series in $N$
\begin{multline}
\tilde{\eta}_{b\to\alpha i}(\mathbf{x}_{\alpha i})=\frac{1}{\Z_{b\to\alpha i}}\prod_{\beta j\in\partial b\backslash\alpha}Tr_{x\beta_{j}}\eta_{\beta j\to b}(\mathbf{x}_{\beta j})\times\\
\exp\left[g(Y_{0},0)\left(1+\frac{1}{N^{(p-1)/2}}S_{b}w_{b}+\frac{1}{N^{p-1}}\left(R_{b}-S_{b}^{2}\right)w_{b}^{2}+\mathcal{O}\text{\ensuremath{\left(\ensuremath{\frac{1}{N^{3(p-1)/2}}}\right)}}\right)\right]\label{eq:tilde_eta_exp}
\end{multline}
where $S_{b}$ and $R_{b}$ are the first and second derivative of
the cost function $g(Y,w)$ evaluated at $Y_{b}$ and $w_{b}=0$:
\begin{align}
S_{b} & \equiv\left.\frac{\partial g(Y_{b},w_{b})}{\partial w}\right|_{w_{b}=0}\\
R_{b} & \equiv\left(\left.\frac{\partial g(Y_{b},w_{b})}{\partial w}\right|_{w_{b}=0}\right)^{2}+\left.\frac{\partial^{2}g(Y_{b},w_{b})}{\partial w_{b}^{2}}\right|_{w_{b}=0}
\end{align}

\subsubsection{Belief propagation}

The mean values of outgoing messages from variable node $\alpha i$
into factor node $a$, are obtained by integrating over the marginal
probabilities $\eta_{\alpha i\to a}$:
\begin{equation}
\mathbf{\hat{x}}_{\alpha i\to i}=\int dx_{\alpha i}\eta_{\alpha i\to a}(\mathbf{x}_{\alpha i})\mathbf{x}_{\alpha i}^{T}\in\R^{r}.
\end{equation}
Note that we have used the transpose of the vector $\mathbf{x}_{\alpha i}^{T}$,
which will become useful for the notation below. Their covariance
matrix is equal to 
\begin{equation}
\hat{\sigma}_{\alpha i\to a}=\int d\mathbf{x}_{\alpha i}\eta_{\alpha i\to a}(\mathbf{x}_{\alpha i})\mathbf{x}_{\alpha i}\mathbf{x}_{\alpha i}^{T}-\hat{\mathbf{x}}_{\alpha i\to i}\hat{\mathbf{x}}_{\alpha)\to i}^{T}\in\R^{r\times r}.
\end{equation}
Using the first- and second-order statistics, we can write explicit
expressions for the moments of $w_{b}$ appearing in the expansion
\eqref{eq:tilde_eta_exp} above. The first moment reads
\begin{multline}
\prod_{\beta j\in\partial b\backslash\alpha}\int d\mathbf{x}_{\beta i}\eta_{\beta)\to b}(\mathbf{x}_{\beta j})w_{b}=\prod_{\beta j\in\partial b\backslash\alpha}\int dx_{\beta i}\eta_{\beta j\to b}(\mathbf{x}_{\beta j})\sum_{\rho=1}^{r}\prod_{\beta j\in\partial b}x_{\beta j}^{\rho}\\
=\mathbf{x}_{\alpha i}^{T}\prod_{\beta j\in\partial b\backslash\alpha}^{\odot}\hat{\mathbf{x}}_{\beta j\to b}.
\end{multline}

Similarly, the second moment, $w_{b}^{2}$, is given by 
\begin{multline}
\prod_{\beta j\in\partial b\setminus\alpha}\int d\mathbf{x}_{\beta i}\eta_{\beta j\to b}(\mathbf{x}_{\beta j})w_{b}^{2}=\int\prod_{\beta j\in\partial b\backslash\alpha}d\mathbf{x}_{\beta i}\eta_{\beta j\to b}(\mathbf{x}_{\beta j})\prod_{\beta j\in\partial b}^{\odot}x_{\beta j}^{\rho T}\prod_{\gamma k\in\partial b}^{\odot}x_{\gamma k}^{\rho}=\\
\mathbf{x}_{\alpha i}^{T}\prod_{\beta j\in\partial b\backslash\alpha}^{\odot}\left(\sigma_{\beta j\to b}+\hat{\mathbf{x}}_{\beta j\to b}\hat{\mathbf{x}}_{\beta j\to b}^{T}\right)\mathbf{x}_{\alpha i}.
\end{multline}
Introducing the explicit moments back into eq. \eqref{eq:tilde_eta_exp},
the incoming messages into variable nodes are given by
\begin{multline}
\tilde{\eta}_{b\to\alpha i}(x_{\alpha i})=\frac{e^{g(Y_{b},0)}}{\Z_{b\to\alpha i}}\left[1+\frac{1}{N^{(p-1)/2}}S_{b}\mathbf{x}_{\alpha i}^{T}\prod_{\beta j\in\partial b\backslash\alpha}^{\odot}\hat{\mathbf{x}}_{\beta j\to b}+\right.\\
\left.\frac{1}{N^{p-1}}\left(R_{b}-S_{b}^{2}\right)x_{\alpha i}^{T}\prod_{\beta j\in\partial b\backslash\alpha}^{\odot}\left(\sigma_{\beta j\to b}+\mathbf{x}_{\beta j\to b}\mathbf{x}_{\beta j\to b}^{T}\right)\mathbf{x}_{\alpha i}\right]\text{+O\ensuremath{\left(\ensuremath{\frac{1}{N^{3(p-1)/2}}}\right)}}.
\end{multline}
Since we are interested in the marginals in the variable nodes, we
can replace this result in the expression for messages outgoing from
a variable node \eqref{eq:outgoing_msg}, we obtain 
\begin{multline}
\eta_{\alpha i\to a}(\mathbf{x}_{\alpha i})=\frac{P_{\alpha}(\mathbf{x}_{\alpha i})}{\Z_{\alpha i\to a}}\prod_{b\in\partial\alpha i\backslash a}\tilde{\eta}_{b\to\alpha i}(\mathbf{x}_{\alpha i})=\frac{P_{\alpha}(\mathbf{x}_{\alpha i})}{\Z_{\alpha i\to a}}\frac{e^{Ng(Y_{b},0)}}{\prod_{b\in\partial\alpha i\backslash a}\Z_{b\to\alpha i}}\times\\
\exp\sum_{b\in\partial\alpha i\backslash a}\left[\frac{1}{N^{p-1}}\left(R_{b}-S_{b}^{2}\right)x_{\alpha i}^{T}\prod_{\beta j\in\partial b\backslash\alpha}^{\odot}\left(\sigma_{\beta j\to b}+\mathbf{x}_{\beta j\to b}\mathbf{x}_{\beta j\to b}^{T}\right)\mathbf{x}_{\alpha i}\right].\label{eq:eta_full}
\end{multline}
Note that $g(Y_{b},0)$ is a constant and can be absorbed into the
normalization function. We define the two order parameters 
\begin{align}
\mathbf{u}_{\alpha i\to a}^{T} & =\frac{1}{N^{(p-1)/2}}\sum_{b\in\partial\alpha i\backslash a}S_{b}\prod_{\beta j\in\partial b\backslash\alpha i}^{\odot}\mathbf{\hat{x}}_{\beta j\to b}^{T}\in\R^{r},\label{eq:stat_u}
\end{align}
and 
\begin{multline}
A_{\beta j\to b}=\frac{1}{N^{p-1}}\sum_{b\in\partial\alpha i\backslash a}\left[\prod_{\beta j\in\partial b\backslash\alpha i}^{\odot}S_{b}^{2}\mathbf{x}_{\beta j\to b}\mathbf{x}_{\beta j\to b}^{T}\right.\\
\left.-R_{b}\prod_{\beta j\in\partial b\backslash\alpha i}^{\odot}\left(\sigma_{\beta j\to b}+\mathbf{x}_{\beta j\to b}\mathbf{x}_{\beta j\to b}^{T}\right)\right]\in\R^{r\times r}.\label{eq:stat_A}
\end{multline}
Using the order parameters we rewrite equation \eqref{eq:eta_full}
as
\begin{equation}
\eta_{\alpha i\to a}(\mathbf{x}_{\alpha i})=\frac{P_{\alpha}(\mathbf{x}_{\alpha i})}{\Z_{\alpha i\to a}}\prod_{b\in\partial\alpha i\backslash a}\exp\left(-\mathbf{x}_{\alpha i}^{T}A_{\beta j\to b}\mathbf{x}_{\alpha i}+\mathbf{u}_{\beta j\to b}^{T}\mathbf{x}_{\alpha i}\right).
\end{equation}
The normalization, or partition function $\Z_{\alpha i\to a}$ , can
be written in terms of the order parameters $\mathbf{u}_{\alpha i\to a}^{T}$
and $A_{\beta j\to b}$ as 
\begin{equation}
\Z_{\alpha i\to a}=Tr_{x_{\alpha i}}P_{\alpha}(\mathbf{x}_{\alpha i})\prod_{b\in\partial\alpha i\backslash a}\exp\left(-\mathbf{x}_{\alpha i}^{T}A_{\beta j\to b}\mathbf{x}_{\alpha i}+\mathbf{u}_{\beta j\to b}^{T}\mathbf{x}_{\alpha i}\right).
\end{equation}
Finally, the moments of the local variables $x_{\alpha i}$ with distribution
$\eta_{\alpha i\to a}(\mathbf{x}_{\alpha i})$ can be found directly
from the partition functions $\Z_{\alpha i\to a}$ by standard derivations.
The mean is given by

\begin{equation}
\hat{x}_{\alpha i\to a}=\frac{\partial}{\partial\mathbf{u}_{\alpha i\to a}}\Z_{\alpha i\to a}(A_{\alpha i\to a},\mathbf{u}_{\alpha i\to a})\equiv f\left(A_{\alpha i\to a},\mathbf{u}_{\alpha i\to a}\right),\label{eq:xhat_alg}
\end{equation}
and the covariance matrices are
\begin{gather}
\sigma_{\alpha i\to a}=\frac{\partial^{2}}{\partial\mathbf{u}_{\alpha i\to a}\partial\mathbf{u}_{\alpha i\to a}^{T}}\Z_{\alpha i\to a}(A_{\alpha i\to a},\mathbf{u}_{\alpha i\to a})\nonumber \\
=\frac{\partial}{\partial\mathbf{u}_{\alpha i\to a}}f\left(A_{\alpha i\to a},\mathbf{u}_{\alpha i\to a}\right).\label{eq:sig_alg}
\end{gather}

\subsection{AMP algorithms}

The mean-field equations, describing the equilibrium of the local
estimators can be used to iteratively into an algorithm by iteratively
calculating the statistics of the messages given their estimators
using eq. \eqref{eq:stat_u} and \eqref{eq:stat_A} and then reevaluating
the estimators $\hat{x}$ and $\sigma$ using eq. \eqref{eq:xhat_alg}
and \eqref{eq:sig_alg}. Defining the upper-script $t$ denoting the
time step of the algorithm iteration, we can write the iterative equations
as 

\begin{align}
\mathbf{u}_{\alpha i\to a}^{t} & =\frac{1}{N^{(p-1)/2}}\sum_{b\in\partial\alpha i\backslash a}S_{b}\prod_{\beta j\in\partial b\backslash\alpha i}^{\odot}\hat{x}_{\beta j\to b}^{t}\label{eq:alg_u1}\\
A_{\alpha i\to a}^{t} & =\frac{1}{N^{p-1}}\sum_{b\in\partial\alpha i\backslash a}\left[S_{b}^{2}\prod_{\beta j\in\partial b\backslash\alpha i}^{\odot}\hat{\mathbf{x}}_{\beta j\to b}^{t}\hat{\mathbf{x}}_{\beta j\to b}^{tT}\right.\\
 & \;-\left.R_{b}\prod_{\beta j\in\partial b\backslash\alpha i}^{\odot}\left(\sigma_{\beta j\to b}^{t}+\hat{\mathbf{x}}_{\beta j\to b}^{t}\hat{\mathbf{x}}_{\beta j\to b}^{tT}\right)\right]\\
\hat{x}_{\alpha i\to a}^{t+1} & =\frac{\partial}{\partial\mathbf{u}_{\alpha i\to a}^{t}}\log\Z_{\alpha i\to a}(A_{\alpha i\to a}^{t},\mathbf{u}_{\alpha i\to a}^{t})\\
\sigma_{\alpha i\to a}^{t+1} & =\frac{\partial^{2}}{\partial\mathbf{u}_{\alpha i\to a}^{t}\partial\mathbf{u}_{\alpha i\to a}^{tT}}\log\Z_{\alpha i\to a}(A_{\alpha i\to a}^{t},\mathbf{u}_{\alpha i\to a}^{t})\label{eq:alg_sig1}
\end{align}

\subsection{Approximate message passing -- local mean-field approximation for
the messages}

In the equations above \eqref{eq:alg_u1}-\eqref{eq:alg_sig1}, the
number of overall messages (and thus calculations) scale with the
number of edges in the factorized graph, i.e. as $\mathcal{O}(N^{P})$.
However, the dependence of each message on the state of the target
node is weak. Therefore, the values of $\mathbf{u}_{\alpha i\to a}^{t}$
and $A_{\alpha i\to a}^{t}$are very close to their mean, when marginalized
over all target nodes $a$. The local deviations about that mean scale
as $N^{(1-p)/2}$. For that reason, we can consider the statistics
of all outgoing messages from each node (i.e., average over all the
adjacent edges), and assume small fluctuations due to the state of
the targets. This procedure is essentially performing mean-field approximation
at every node. The result will be the AMP equations which scale with
the number of variable nodes $PN,$ rather than with the number of
edges in the graph. In physics, this analogous to the cavity method.

To apply this reasoning to the equations, we define the order parameters
$A_{\alpha i}$ and $\mathbf{u}_{\alpha i}$, which explicitly exclude
the dependence of the target node:
\begin{align}
\mathbf{u}_{\alpha i}^{t} & =\frac{1}{N^{(p-1)/2}}\sum_{b\in\partial\alpha i}S_{b}\prod_{\beta j\in\partial b}^{\odot}\hat{\mathbf{x}}_{\beta j\to b}^{t},\label{eq:AMP_u_def}\\
A_{\alpha i}^{t} & =\frac{1}{N^{p-1}}\sum_{b\in\partial\alpha i}\left[S_{b}^{2}\prod_{\beta j\in\partial b\backslash\alpha i}^{\odot}\hat{\mathbf{x}}_{\beta j\to b}^{t}\hat{\mathbf{x}}_{\beta j\to b}^{tT}-R_{b}\prod_{\beta j\in\partial b\backslash\alpha i}^{\odot}\left(\sigma_{\beta j\to b}^{t}+\hat{\mathbf{x}}_{\beta j\to b}^{t}\mathbf{\hat{x}}_{\beta j\to b}^{tT}\right)\right].
\end{align}
The difference between the non-directed and the directed messages
is the component that depends on the target node $S_{a}$. For the
mean-messages, the correction terms scale as $O(N^{(1-p)/2})$, and
is given by 
\begin{equation}
\delta\mathbf{u}_{\alpha i\to a}^{t}=\mathbf{u}_{\alpha i}^{t}-\mathbf{u}_{\alpha i\to a}^{t}=\frac{1}{N^{(p-1)/2}}S_{a}\prod_{\beta j\in\partial a\backslash\alpha i}^{\odot}\hat{\mathbf{x}}_{\beta j\to a}^{t}.\label{eq:AMP_u}
\end{equation}
For the fluctuations in the local messages about their mean, the correction
term scales as 
\begin{equation}
A_{\alpha i}^{t}-A_{\alpha i\to a}^{t}\sim\mathcal{O}(N^{1-p}),
\end{equation}
and we will be neglecting it. 

To transform the equations for the local messages statistics, to use
only the \emph{target-agnostic} variables, $\hat{x}_{\beta j}^{t}$
and $\sigma_{\beta j}^{t}$, we calculate the difference between the
two mean values 
\begin{equation}
\delta\hat{x}_{\alpha i\to a}=\mathbf{\hat{x}}_{\alpha i}^{t}-\hat{\mathbf{x}}_{(\alpha i)\to a}^{t}=f\left(A_{\alpha i\to a}^{t-1},\mathbf{u}_{\alpha i\to a}^{t-1}\right)-f\left(A_{\alpha i}^{t-1},\mathbf{u}_{\alpha i}^{t-1}\right).
\end{equation}
We develop the second term on the RHS to linear order in the small
parameter of the difference $\delta\mathbf{u}_{\alpha i\to a}$, and
note that the leading order cancel with the first term in the RHS
above, yielding
\begin{multline}
\delta\hat{x}_{\alpha i\to a}^{t}=f\left(A_{\alpha i}^{t-1},\mathbf{u}_{\alpha i}^{t-1}\right)+\frac{\partial}{\partial\mathbf{u}}f\left(A_{\alpha i}^{t-1},\mathbf{u}_{\alpha i}^{t-1}\right)\left(\mathbf{u}_{\alpha i\to a}^{t-1}-\mathbf{u}_{\alpha i}^{t-1}\right)-f\left(A_{\alpha i}^{t-1},\mathbf{u}_{\alpha i}^{t-1}\right)=\\
\sigma_{\alpha i}^{t}\left(\mathbf{u}_{\alpha i\to a}^{t-1}-\mathbf{u}_{\alpha i}^{t-1}\right)=\sigma_{\alpha i}^{t}\frac{1}{N^{(p-1)/2}}S_{a}\prod_{\beta j\in\partial a\backslash\alpha i}^{\odot}\hat{\mathbf{x}}_{\beta,j}^{t-1}.\label{eq:AMP_dx}
\end{multline}
Using eq. \eqref{eq:AMP_u} and \eqref{eq:AMP_dx}, we can write an
expression for the node-average local messages, 
\begin{equation}
\mathbf{u}_{\alpha i}^{t}=\frac{1}{N^{(p-1)/2}}\sum_{b\in\partial\alpha i}S_{b}\prod_{\beta j\in\partial b\backslash\alpha i}^{\odot}\left(\mathbf{\hat{x}}_{\beta j}^{t}-\delta\hat{\mathbf{x}}_{\beta j\to b}^{t}\right).\label{eq:u_before_exp}
\end{equation}
Expanding the product of the $\beta j$ factors, and keeping terms
up to linear order in the small difference $\delta x$, we obtain
\begin{equation}
\mathbf{u}_{\alpha i}^{t}=\frac{1}{N^{(p-1)/2}}\sum_{b\in\partial\alpha i}S_{b}\left[\prod_{\beta j\in\partial b\backslash\alpha i}^{\odot}\hat{\mathbf{x}}_{\beta j}^{t}-\sum_{\beta j\in\partial b\backslash\alpha i}\delta\hat{\mathbf{x}}_{\beta j\to b}^{t}\prod_{\gamma k\in\partial b\backslash\alpha i,\beta j}^{\odot}\mathbf{\hat{x}}_{\gamma k}^{t}\right]+O(\frac{1}{N^{(p-1)}}).
\end{equation}
The first correction for the above, involves the quadratic terms in
the expansion of \eqref{eq:u_before_exp}. The mixed terms involve
the values at time $t$ \textbf{and} at\textbf{ t}ime $t-1$, which
originate from the expansion of $\delta x$ in \eqref{eq:AMP_dx}.
The mixed term is given by
\begin{multline}
\frac{1}{N^{(p-1)}}\sum_{b\in\partial\alpha i}S_{b}^{2}\sum_{\beta j\in\partial b\backslash\alpha i}\sigma_{\beta j}^{t}\prod_{\gamma,k\in\partial b\backslash\beta j}^{\odot}\hat{\mathbf{x}}_{\gamma k}^{t-1}\prod_{\gamma k\in\partial b\backslash\alpha i,\beta j}^{\odot}\hat{\mathbf{x}}_{\gamma k}^{t}=\\
\frac{1}{N^{(p-1)}}\hat{x}_{\alpha i}^{t-1}\sum_{b\in\partial\alpha i}S_{b}^{2}\sum_{\beta j\in\partial b\backslash\alpha i}\sigma_{\beta j}^{t}\prod_{\gamma k\in\partial b\backslash\alpha i,\beta j}^{\odot}\hat{\mathbf{x}}_{\gamma k}^{t}\hat{\mathbf{x}}_{\gamma k}^{t-1}.
\end{multline}
This expression, which couples the dynamical variable $\hat{x}$ into
its previous time step is an\emph{ Onsager response term}. It reflects
the changes to the fields of the nodes surrounding the node $\alpha i$
due to the activity of the node $\alpha i$ in the previous time step. 

Importantly, up until this point, we have not yet used the assumption
of Bayes optimality, nor have we used the Nishimori identities that
follow the Bayes-optimal assumption. Consequently, algorithms based
on the approximate message-passing above should be general and do
not require the Bayes-optimal assumption. In the following section,
we consider simplification due to the Bayes-optimal assumption. Beyond
simplification of the mathematical expressions, it will allow us to
systematically derive a dynamical mean-field theory for the errors
in section \ref{sec:Dynamic-mean-field}.

\subsection{Simplifications for Bayes-optimal settings}

The covariance matrix in the Bayes-optimal case can be much simplified.
First, one can show, using the Nishimori identities at the equilibrium,
that 
\begin{equation}
\left\langle R_{b}\right\rangle \equiv\left\langle \left.\frac{\partial g(Y_{b},w)}{\partial w}\right|_{w=0}^{2}\right\rangle +\left\langle \left.\frac{\partial^{2}g(Y_{b},w)}{\partial w^{2}}\right|_{w=0}\right\rangle =0.
\end{equation}
Here the angular brackets denote averaging over the posterior. The
posterior variance of $S_{b}$ is given by the Fisher information
of the output channel $\E_{post}\left[S_{b}^{2}\right]=\frac{1}{\Delta}$.
Furthermore, in Bayes-optimal setting all samples from the equilibrium
ensemble are similar, the sum over nodes becomes self-averaging and
so
\begin{equation}
\sum_{b}S_{b}^{2}=\E_{Post}\left[S_{b}^{2}\right]=\frac{1}{\Delta}.
\end{equation}
Using the above simplifications, the covariance matrix of the messages
can be written as
\begin{equation}
A_{\alpha i}^{t}=\frac{1}{N^{p-1}}\frac{1}{\Delta}\sum_{b\in\partial\alpha i}\prod_{\beta j\in\partial b\backslash\alpha i}^{\odot}\hat{\mathbf{x}}_{\beta j\to b}^{t}\hat{\mathbf{x}}_{\beta j\to b}^{tT}=\frac{1}{\Delta}\prod_{\beta\neq\alpha}^{\odot}\left(\frac{1}{N}\sum_{j=1}^{N}\hat{\mathbf{x}}_{\beta j}^{t}\hat{\mathbf{x}}_{\beta j}^{tT}\right)\equiv A_{\alpha}^{t}.
\end{equation}
Importantly, the covariance matrix does not depend on the specific
node $i$, but it does depend on the mode of the tensor $\alpha$.
This is a significant difference from the algorithms for symmetric-tensor
decomposition, for which $A^{t}$ was uniform for \emph{all} nodes
in the factor graph \cite{lesieur2017statistical}. 

The Onsager term, which appears in the iterative mean-field equations
for the local mean of the messages can also be simplified under the
Bayes-optimal setting. Using some algebra, the Onsager correction
becomes
\begin{multline}
\frac{1}{N^{(p-1)}}\hat{\mathbf{x}}_{\alpha i}^{t-1}\sum_{b\in\partial\alpha i}S_{b}^{2}\sum_{\beta j\in\partial b\backslash\alpha i}\sigma_{\beta j}^{t}\prod_{(\gamma,k)\in\partial b\backslash\alpha i,(\beta j)}^{\odot}\hat{\mathbf{x}}_{\gamma k}^{t}\mathbf{\hat{x}}_{\gamma k}^{t-1}=\\
\frac{1}{\Delta N}\hat{\mathbf{x}}_{\alpha i}^{t-1}\sum_{b\in\partial\alpha i}\sum_{\beta j\in\partial b\backslash\alpha i}\sigma_{\beta j}^{t}\prod_{\gamma\neq\alpha,\beta}^{\odot}\left(\frac{1}{N}\sum_{k}\hat{\mathbf{x}}_{\gamma k}^{t}\hat{\mathbf{x}}_{\gamma k}^{t-1}\right)=\\
\frac{1}{\Delta N}\hat{\mathbf{x}}_{\alpha i}^{t-1}\sum_{\beta\neq\alpha}\sum_{j}\sigma_{\beta j}^{t}\odot\prod_{\gamma\neq\alpha,\beta}^{\odot}\left(\frac{1}{N}\sum_{k}\hat{\mathbf{x}}_{\gamma k}^{t}\hat{\mathbf{x}}_{\gamma k}^{t-1}\right)\equiv\\
\frac{1}{\Delta N}\hat{\mathbf{x}}_{\alpha i}^{t-1}\sum_{\beta\neq\alpha}\sum_{j}\sigma_{\beta j}^{t}\odot D_{\alpha\beta}^{t},
\end{multline}
where 
\begin{equation}
D_{\alpha\beta}^{t}=\prod_{\gamma\neq\alpha,\beta}^{\odot}\left(\frac{1}{N}\sum_{k}\hat{\mathbf{x}}_{\gamma k}^{t}\hat{\mathbf{x}}_{\gamma k}^{t-1}\right).
\end{equation}
Finally, we write the simplified AMP equations as
\begin{align}
\mathbf{u}_{\alpha i}^{t} & =\frac{1}{N^{(p-1)/2}}\sum_{b\in\partial\alpha i}S_{b}\prod_{\beta j\in\partial b\backslash\alpha i}^{\odot}\hat{\mathbf{x}}_{\beta j}^{t}-\frac{1}{\Delta}\mathbf{\hat{x}}_{\alpha i}^{t-1}\sum_{\beta\neq\alpha}\Sigma_{\beta}^{t}\odot D_{\alpha\beta}^{t}\label{eq:AMP_alg_u}\\
A_{\alpha}^{t} & =\frac{1}{\Delta}\prod_{\beta\neq\alpha}^{\odot}\left(\frac{1}{N}\sum_{j=1}^{N}\hat{\mathbf{x}}_{\beta j}^{t}\hat{\mathbf{x}}_{\beta j}^{tT}\right)\\
\mathbf{\hat{x}}_{\alpha i}^{t+1} & =\frac{\partial}{\partial\mathbf{u}_{\alpha i}^{t}}\log\Z_{\alpha}(A_{\alpha}^{t},\mathbf{u}_{\alpha i}^{t})\label{eq:AMP_alg_x}\\
\sigma_{\alpha i}^{t+1} & =\frac{\partial^{2}}{\partial\mathbf{u}_{\alpha i}^{t}\partial\mathbf{u}_{\alpha i}^{tT}}\log\Z_{\alpha}(A_{\alpha}^{t},\mathbf{u}_{\alpha i}^{t}),\label{eq:AMP_alg_sig}
\end{align}
where the Onsager term is given by
\begin{align}
D_{\alpha\beta}^{t} & =\prod_{\gamma\neq\alpha,\beta}^{\odot}\left(\frac{1}{N}\sum_{k}\hat{\mathbf{x}}_{\gamma k}^{t}\hat{\mathbf{x}}_{\gamma k}^{t-1}\right)\\
\Sigma_{\alpha}^{t} & =N^{-1}\sum_{i}\sigma_{\alpha_{i}}^{t},
\end{align}
and the partition function reads 
\begin{equation}
\Z_{\alpha}(A_{\alpha}^{t},\mathbf{u}_{\alpha i}^{t})=\intop d\mathbf{x}P_{\alpha}(\mathbf{x})\exp\left[\left(\mathbf{u}_{\alpha i}^{T}x-\mathbf{x}^{T}A_{\alpha}^{t}\mathbf{x}\right)\right].
\end{equation}

There are two parameters in these equations (apart from the prior
distributions $P_{\alpha}(x)$. One is the Fisher information of the
output channel, which is a global parameter that can tune the global
dynamics. The other $S_{b}$, which is the Fisher score of the entry
at $Y_{b}$. The last one is what yields the structure in the solution
of the estimators.

\section{Dynamic mean field theory (state evolution) \label{sec:Dynamic-mean-field}}

In the previous section we have derived the AMP algorithm for general
tensors, and show their simplified form in the case of the Bayes-optimal
assumption, where the priors are known, and the system follow Nishimori
identities at equilibrium. These algorithms follow the iterative evolution
of the estimators in each of the variable nodes in the factor graph.
In order to analytically study the performance of the algorithm, we
want to know how the mean error reduces from one iteration of the
algorithm to the next. To do that, we derive a dynamical mean-field
theory (also known as \emph{state-evolution}). As mentioned above,
following the Bayes-optimal assumption, the estimators are self-averaging;
thus a mean-field description of the error is a good measure for the
typical evolution of any given system.

We define an order parameter that measures the overlap between each
of the underlying vectors of estimators $\hat{x}_{\alpha}^{t}\in\mathbb{R}^{p\times r}$
and the ground truth values $x_{\alpha}^{0}\in\mathbb{R}^{p\times r}$.
The \emph{overlap matrix} is defined as
\begin{equation}
M_{\alpha}^{t}=\frac{1}{N}\sum_{i}^{N}\hat{\mathbf{x}}_{\alpha i}^{t}\mathbf{x}_{\alpha i}^{0T}\in\R^{r\times r}.\label{eq:DMF_M_def}
\end{equation}
In total, there are $p$ matrices of dimensions $r\times r$, each
for each mode of the tensor. In the Bayes-optimal regime, the ground-truth
values can be replaced with any typical sample from the posterior
distribution. Thus, in Bayes-optimal inference, $M_{\alpha}^{t}$
is also the typical covariance matrix of the estimators

\begin{equation}
\frac{1}{N}\sum_{i}^{N}\hat{\mathbf{x}}_{\alpha i}^{t}\hat{\mathbf{x}}_{\alpha i}^{t}=M_{\alpha}^{t}.
\end{equation}
It follows that under the Bayes-optimality condition, $M_{\alpha}^{t}$
is a symmetric matrix.

To study the typical dynamics of the algorithm using the mean overlap,
we derive yet another mean-field approximation, now on the spatial
degrees of freedom -- i.e., the nodes. Given the self-averaging property
of the nodes under the Bayes-optimal setting, and using the central-limit
theorem, we need to find the first two moments of the distribution
of the local values $\mathbf{u}_{i\alpha}$ (note that $A_{\alpha}$
are already node-independent). Following the usual procedure of mean-field
theory, we then close the equations self-consistently using the overlap
parameter $M_{\alpha}^{t}$. 

Using the definition of $\mathbf{u}_{\alpha i}$ from eq. \eqref{eq:AMP_u_def},
we average over the posterior $P_{out}$:
\begin{multline}
\E\left[\mathbf{u}_{\alpha i}^{t}\right]=\frac{1}{N^{(p-1)/2}}\sum_{b\in\partial\alpha i}\E\left[S_{b}\prod_{\beta j\in\partial b\backslash\alpha i}^{\odot}\hat{\mathbf{x}}_{\beta j\to b}^{t}\right]=\\
\frac{1}{N^{(p-1)/2}}\sum_{b\in\partial\alpha i}\intop dY_{b}P_{out}(Y_{b},w_{b})\left.\frac{\partial\log P_{out}\cond{Y_{b}}w}{\partial w}\right|_{w=0}\prod_{\beta j\in\partial b\backslash\alpha i}^{\odot}\hat{\mathbf{x}}_{\beta j\to b}^{t}\label{eq:dmf_exp_u}
\end{multline}
Similar to the the approximation carried above for the AMP algorithms,
we develop the posterior probability about $w=0$, and keep only the
leading terms,
\begin{equation}
P_{out}(Y_{b},w_{b})=P_{out}(Y_{b},0)+P_{out}(Y_{b},0)w_{b}\left(\frac{\partial\log P_{out}(Y_{b},w)}{\partial w}\right)_{w=0}+O(w^{2}).
\end{equation}
Carrying the integration in eq. \eqref{eq:dmf_exp_u}, the leading
order will vanish 
\begin{equation}
\intop dY_{b}P_{out}(Y_{b},0)\left.\frac{\partial\log P_{out}\cond{Y_{b}}w}{\partial w}\right|_{w=0}=0,
\end{equation}
which is the consequence of the Nishimori identities. In other words,
in a Bayes-optimal setting, and when the interactions are weak, then
the average value of the messages when averaged over the entire graph
vanish to leading order. Intuitively, since the underlying graph is
isotropic, we expect that the dynamics will be similar at every node
on average.

Performing the integration on the next, quadratic, term in \eqref{eq:dmf_exp_u}
we get
\begin{multline}
\E\left[\mathbf{u}_{\alpha i}^{t}\right]=\frac{1}{N^{(p-1)/2}}\times\\
\sum_{b\in\partial\alpha i}P_{out}(Y_{b},0)w_{b}\left(\frac{\partial\log P_{out}(Y_{b},w)}{\partial w}\right)_{w=0}^{2}\prod_{\beta j\in\partial b\backslash\alpha i}^{\odot}\hat{\mathbf{x}}_{\beta j\to b}^{t}=\\
\frac{1}{\Delta N^{(p-1)/2}}\sum_{b\in\partial\alpha i}w_{b}\prod_{\beta j\in\partial b\backslash\alpha i}^{\odot}\mathbf{\hat{x}}_{\beta j\to b}^{t}.
\end{multline}
Note that the original tensor components, denoted by $w_{a}$ are
the ground-truth in the context of the inference problem, and we can
write 
\begin{equation}
w_{a}=\frac{1}{N^{\frac{p-1}{2}}}\prod_{(\beta j)\in\partial a}^{\odot}\boldsymbol{x}_{\beta j}^{0}.
\end{equation}
 Replacing this into the expression for the expectation above we get
\begin{multline}
\E\left[\mathbf{u}_{\alpha i}^{t}\right]=\frac{1}{\Delta N^{(p-1)}}\sum_{b\in\partial\alpha i}\sum_{\rho=1}^{r}\prod_{(\beta j)\in\partial b}^{\odot}\mathbf{x}_{\beta j}^{0,\rho}\prod_{\beta j\in\partial b\backslash\alpha i}^{\odot}\hat{\mathbf{x}}_{\beta j}^{tT}\\
=\frac{1}{\Delta N^{(p-1)}}\sum_{b\in\partial\alpha i}\left(\mathbf{x}_{\alpha i}^{0}\right)^{T}\left(\prod_{\beta j\in\partial b\backslash\alpha i}^{\odot}\mathbf{x}_{\beta j}^{0}\right)\left(\prod_{\beta j\in\partial b\backslash\alpha i}^{\odot}\hat{\mathbf{x}}_{\beta j}^{t}\right)^{T}\\
=\frac{1}{\Delta}\left(\mathbf{x}_{\alpha i}^{0}\right)^{T}\prod_{\beta\neq\alpha}^{\odot}\left(\frac{1}{N}\sum_{j}^{N}\mathbf{x}_{\beta j}^{0}\hat{\mathbf{x}}_{\beta j}^{tT}\right)=\frac{1}{\Delta}\left(\mathbf{x}_{\alpha i}^{0}\right)^{T}\prod_{\beta\neq\alpha}^{\odot}M_{\beta}^{t}.
\end{multline}
Finally we can write
\begin{equation}
\E\left[\mathbf{u}_{\alpha i}^{t}\right]=\frac{1}{\Delta}\prod_{\beta\neq\alpha}^{\odot}M_{\beta}^{t}\mathbf{x}_{\alpha i}^{0}.\label{eq:dmf_exp_u-M}
\end{equation}
In a similar manner, we can calculate the covariance matrix of the
mean-messages 
\begin{equation}
cov\left[\mathbf{u}_{i\alpha}^{t}\right]=\sum_{a\in\partial\alpha i}\intop dY_{a}P_{out}(Y_{a},w)\left(\frac{\partial g(Y_{a},w)}{\partial w}\right)_{w=0}^{2}\frac{1}{N^{p-1}}\prod_{\beta j\in\partial a\backslash\alpha i}^{\odot}\mathbf{\hat{x}}_{\beta j}^{t}\hat{\mathbf{x}}_{\beta j}^{tT}.
\end{equation}
Keeping the leading order after the expansion of the distribution
$P_{out}$ for small $w$ we get 
\begin{equation}
cov\left[\mathbf{\mathbf{u}}_{i\alpha}^{t}\right]=\frac{1}{N^{p-1}\Delta}\sum_{a\in\partial\alpha i}\prod_{\beta j\in\partial a\backslash\alpha i}^{\odot}\hat{\mathbf{x}}_{\beta j}^{t}\hat{\mathbf{x}}_{\beta j}^{tT}.
\end{equation}
In the Bayes-optimal setting this is equal to 
\begin{equation}
cov\left[\mathbf{u}_{i\alpha}^{t}\right]=\frac{1}{\Delta}\prod_{\beta\neq\alpha}^{\odot}M_{\beta}^{t}.
\end{equation}
While the mean-message $\mathbf{u}_{\alpha i}$ varies from node to
node, the mean covariance (not to be confused with the \emph{covariance
of the mean} calculated above), $A_{\alpha}^{t}$, is node-independent,
as we have established in the previous section. In the Bayes-optimal
setting, where the Nishimori identities hold, it is equal to
\begin{equation}
A_{\alpha}^{t}=\frac{1}{\Delta}\prod_{\beta\neq\alpha}^{\odot}\left(\frac{1}{N}\sum_{j=1}^{N}\hat{\mathbf{x}}_{\beta j}^{t}\mathbf{\hat{x}}_{\beta j}^{tT}\right)=\frac{1}{\Delta}\prod_{\beta\neq\alpha}^{\odot}M_{\beta}^{t}.
\end{equation}
Using the definition of the mean overlap in eq. \eqref{eq:DMF_M_def},
and eq. \eqref{eq:AMP_alg_x}, we write 
\begin{equation}
M_{\alpha}^{t}=\frac{1}{N}\sum_{i}^{N}\hat{\mathbf{x}}_{\alpha i}^{t}\mathbf{x}_{\alpha i}^{0T}=\frac{1}{N}\sum_{i}f_{\alpha}\left(A_{\alpha}^{t-1},\mathbf{u}_{\alpha i}^{t-1}\right)\mathbf{x}_{\alpha i}^{0T},\label{eq:DMF_M_1}
\end{equation}
where 
\begin{equation}
f_{\alpha}\equiv\frac{\partial}{\partial\mathbf{u}}\log\Z(A_{\alpha},\mathbf{u}_{\alpha}),
\end{equation}
and with the partition function 
\begin{equation}
\Z_{\alpha}(A_{\alpha},\mathbf{u}_{\alpha})=\intop d\mathbf{x}P_{\alpha}(\mathbf{x})\exp\left[\left(\mathbf{u_{\alpha}}^{T}\mathbf{x}-x^{T}A_{\alpha}^{t}\mathbf{x}\right)\right].
\end{equation}
Replacing the average over all nodes $i$ in \eqref{eq:DMF_M_1} with
the expectation, we write an iterative update equation for the order
parameter $M_{\alpha}^{t}$,
\begin{equation}
M_{\alpha}^{t+1}=\int d\boldsymbol{x}_{\alpha}^{0}P_{\alpha}(\boldsymbol{x}_{\alpha}^{0})\E_{z}\left[f_{\alpha}\left(\frac{1}{\Delta}\prod_{\beta\neq\alpha}^{\odot}M_{\beta}^{t},\frac{1}{\Delta}\prod_{\beta\neq\alpha}^{\odot}M_{\beta}^{t}x_{\alpha i}^{0}+\frac{1}{\sqrt{\Delta}}\left(\prod_{\beta\neq\alpha}^{\odot}M_{\beta}^{t}\right)^{\frac{1}{2}}z\right)\mathbf{x}_{\alpha i}^{0T}\right].\label{eq:DMFT_SE-1}
\end{equation}
Here, $z\in\R^{r}$ are random variables with standard normal distribution.
The expectation in the RHS of eq. \eqref{eq:DMFT_SE-1} is over two
random variables: First are expected values for the underlying ground-truth
$\boldsymbol{x}_{\alpha}^{0}$, which follows the prior distribution
$P_{\alpha}$; The second is of a standard gaussian variable $z$,
which represent the node-to-node fluctuations in the local mean-messages,
with mean $\frac{1}{\Delta}\prod_{\beta\neq\alpha}^{\odot}M_{\beta}^{t}x_{\alpha i}^{0}$
and covariance matrix $\frac{1}{\Delta}\prod_{\beta\neq\alpha}^{\odot}M_{\beta}^{t}$.

The final overlap values of the iterative algorithms are given by
the stable fixed points of the dynamic equations defined in \eqref{eq:DMFT_SE-1}.
These can be obtained by finding the solutions $M_{\alpha}^{*}$ for
the $p$ equations 
\begin{equation}
M_{\alpha}^{*}=\int d\boldsymbol{x}_{\alpha}^{0}P_{\alpha}(\boldsymbol{x}_{\alpha}^{0})\E_{z}\left[f_{\alpha}\left(\frac{1}{\Delta}\prod_{\beta\neq\alpha}^{\odot}M_{\beta}^{*},\frac{1}{\Delta}\prod_{\beta\neq\alpha}^{\odot}M_{\beta}^{*}x_{\alpha i}^{0}+\frac{1}{\sqrt{\Delta}}\left(\prod_{\beta\neq\alpha}^{\odot}M_{\beta}^{*}\right)^{\frac{1}{2}}z\right)\mathbf{x}_{\alpha i}^{0T}\right].
\end{equation}

\paragraph{Mean square error}

The real quantity of interest is the mean square error (MSE) of the
estimate. This can be easily obtained from the mean overlap at any
time-step of the algorithm using.
\begin{equation}
MSE_{\alpha}^{t}=\frac{1}{\sigma_{\alpha}^{2}}Tr\left[\E_{P_{\alpha}}\left[\mathbf{x^{0}}_{\alpha}\mathbf{x}_{\alpha}^{0T}\right]-M_{\alpha}^{t}\right].
\end{equation}

Finally, the expected error of the AMP algorithms, once it has converged
is given by 
\begin{equation}
MSE_{\alpha}^{AMP}=\frac{1}{\sigma_{\alpha}^{2}}Tr\left[\E_{P_{\alpha}}\left[\mathbf{x^{0}}_{\alpha}\mathbf{x}_{\alpha}^{0T}\right]-M_{\alpha}^{*}\right].
\end{equation}

\section{Convergence of the AMP algorithms}

Approximate message passing, and belief-propagation algorithms in
general are known to have convergence issues (see for example \cite{pretti2005message,bickson2008gaussian,johnson2009fixing,rangan2014convergence,manoel2015swept}).
A typical naive implementation of the algorithms will reduce the overall
mean square error of the estimator, $MSE^{t}$. However, at some point,
$MSE^{t}$ will start increasing and may diverge to large deviations
from the ground-truth values or will oscillate about some fixed value.
Loosely speaking, the step size of the iterative update equations
\eqref{eq:AMP_alg_x} and \eqref{eq:AMP_alg_sig} is too big, and
the algorithm may 'overshoot' the MMSE estimator. One possible way
to correct this behavior (see e.g., \cite{rangan2014convergence}
and reference therein) is to reduce the step size. Since the differential
change to $x^{t}$ and $\sigma^{t}$ is proportional to derivatives
of the partition function in \eqref{eq:AMP_alg_x} and \eqref{eq:AMP_alg_sig},
a good normalization scheme could use an energy estimation of the
configuration at time-step $t$. To do this, one can evaluate the
Bethe free energy at every time step \cite{rangan2014convergence,manoel2015swept,lesieur2017constrained}.
However, since this report does not focus on possible implementations
of the algorithms, it is sufficient to use a simpler -- and potentially
less efficient -- scheme, using fixed step-size reduction, or \emph{damping}. 

To implement the fixed damping algorithm, eq \eqref{eq:AMP_alg_x}
and \eqref{eq:AMP_alg_sig} can be rewritten as 
\begin{align}
\mathbf{\hat{x}}_{\alpha i}^{t+1} & =\lambda\mathbf{\hat{x}}_{\alpha i}^{t+1}+(1-\lambda)\frac{\partial}{\partial\mathbf{u}_{\alpha i}^{t}}\log\Z_{\alpha}(A_{\alpha}^{t},\mathbf{u}_{\alpha i}^{t})\\
\sigma_{\alpha i}^{t+1} & =\lambda\sigma_{\alpha i}^{t+1}+(1-\lambda)\frac{\partial^{2}}{\partial\mathbf{u}_{\alpha i}^{t}\partial\mathbf{u}_{\alpha i}^{tT}}\log\Z_{\alpha}(A_{\alpha}^{t},\mathbf{u}_{\alpha i}^{t})
\end{align}
where $0\leq\lambda<1$ is the damping coefficient that controls the
effective step size and the speed of convergence. In this simple implementation,
the level of damping is a control parameter of the algorithm. A more
sophisticated approach would use adaptive damping$\lambda_{t}$, where
the effective step size decreases as the Bethe free energy of the
configuration $\{\boldsymbol{x}_{\alpha}^{t}\}$ decreases \cite{pretti2005message,lesieur2017constrained}.

\section{Noncubic tensors }

In the above, we have assumed that the dimensionality of all $p$
modes is $N$, implying that the underlying tensor is cubic (i.e.,
all modes have the same length). To study how the shape of the tensors
influence the AMP algorithm and the performance, we allow for the
different modes to have different dimensionality $N_{\alpha}$. Importantly,
we assume that all modes are in the thermodynamic regime, i.e., $N_{\alpha}\to\infty$
$\alpha=\{1,...,p\}$. Furthermore, we assume all modes scale in a
similar way. This is done by defining $N_{\alpha}=n_{\alpha}N$ where
all $n_{\alpha}=O(1)$ and $\prod_{\alpha}n_{\alpha}=1$. The thermodynamic
limit is then understood by taking $N\to\infty$. 

First we note that the scaling of the tensor elements does not change
with this choice of scaling,
\[
w_{b}\sim\sqrt{\frac{N}{\prod_{\alpha}N_{\alpha}}}\sim\frac{N^{-\frac{p-1}{2}}}{\sqrt{\prod_{\alpha}n_{\alpha}}}=N^{\frac{1-p}{2}}.
\]
However, the algorithms have no symmetry with respect to the dimensionality
of the different modes in this case. This broken symmetry is in the
iterative mean-field equations for the local mean messages $\mathbf{u}_{\alpha i}$,
in eq. \eqref{eq:AMP_alg_u} , which now is scaled by proportion of
the dimensionality respective mode:
\begin{equation}
\mathbf{u}_{\alpha i}^{t}=\frac{n_{\alpha}}{N^{(p-1)/2}}\sum_{b\in\partial\alpha i}S_{b}\prod_{\beta j\in\partial b\backslash\alpha i}^{\odot}\hat{\mathbf{x}}_{\beta j}^{t}-\frac{1}{\Delta}\hat{\mathbf{x}}_{\alpha i}^{t-1}\sum_{\beta\neq\alpha}\Sigma_{\beta}^{t}\odot D_{\alpha\beta}^{t}
\end{equation}
The other mean-field equations of the algorithms are left unchanged.

\paragraph{Correction to the dynamic mean-filed equations}

In order to make the necessary changes to the dynamic mean-field theory
in section \ref{sec:Dynamic-mean-field}, we redefine the mean overlap
with the appropriate scaling, which now depends on the mode $\alpha$,
\begin{equation}
M_{\alpha}^{t}=\frac{1}{n_{\alpha}N}\sum_{i}^{N}\hat{\mathbf{x}}_{\alpha i}^{t}\mathbf{x}_{\alpha i}^{0T}\in\R^{r\times r}.
\end{equation}

Using the rescaled overlap, we re-derive the iterative dynamic mean-field
equations, following the same steps as in section \ref{sec:Dynamic-mean-field}:

\begin{multline}
\E\left[\mathbf{u}_{\alpha i}^{t}\right]=\frac{n_{\alpha}}{N^{(p-1)/2}\sqrt{\prod_{\beta}n_{\beta}}}\times\\
\sum_{b\in\partial\alpha i}P_{out}(Y_{b},0)w_{b}\left(\frac{\partial\log P_{out}(Y_{b},w)}{\partial w}\right)_{w=0}^{2}\prod_{\beta j\in\partial b\backslash\alpha i}^{\odot}\hat{\mathbf{x}}_{\beta j\to b}^{t}=\\
\frac{1}{\Delta N^{(p-1)/2}}\sum_{b\in\partial\alpha i}w_{b}\prod_{\beta j\in\partial b\backslash\alpha i}^{\odot}\hat{\mathbf{x}}_{\beta j\to b}^{t}.
\end{multline}
Substituting the expression for $w$,
\begin{multline}
\E\left[\mathbf{u}_{\alpha i}^{t}\right]=\frac{n_{\alpha}}{\Delta N^{(p-1)}\prod_{\beta}n_{\beta}}\sum_{b\in\partial\alpha i}\prod_{(\beta j)\in\partial b}^{\odot}\mathbf{x}_{\beta j}^{0}\prod_{\beta j\in\partial b\backslash\alpha i}^{\odot}\hat{x}_{\beta j}^{tT}\\
=\frac{n_{\alpha}}{\Delta N^{(p-1)}\prod_{\beta}n_{\beta}}\sum_{b\in\partial\alpha i}\left(\mathbf{x}_{\alpha i}^{0}\right)^{T}\left(\prod_{\beta j\in\partial b\backslash\alpha i}^{\odot}\mathbf{x}_{\beta j}^{0}\right)\left(\prod_{\beta j\in\partial b\backslash\alpha i}^{\odot}\hat{\mathbf{x}}_{\beta j}^{t}\right)^{T}\\
=\frac{n_{\alpha}}{\Delta}\left(\mathbf{x}_{\alpha i}^{0}\right)^{T}\prod_{\beta\neq\alpha}^{\odot}\left(\frac{1}{Nn_{\beta}}\sum_{j}^{N}\mathbf{x}_{\beta j}^{0}\mathbf{\hat{x}}_{\beta j}^{tT}\right)=\frac{n_{\alpha}}{\Delta}\left(\mathbf{x}_{\alpha i}^{0}\right)^{T}\prod_{\beta\neq\alpha}^{\odot}M_{\beta}^{t}
\end{multline}
Finally we arrive at
\begin{equation}
\E\left[\mathbf{u}_{\alpha i}^{t}\right]=\frac{n_{\alpha}}{\Delta}\prod_{\beta\neq\alpha}^{\odot}M_{\beta}^{t}\mathbf{x}_{\alpha i}^{0}.
\end{equation}
Note that the only difference between this result and eq. \eqref{eq:dmf_exp_u-M}
is the factor $n_{\alpha}.$ It follows that in order to generalize
the dynamic mean-field theory to noncubic tensors, we simply need
to replace $M_{\alpha}\to n_{\alpha}M_{\alpha}$ throughout the results
of section \ref{sec:Dynamic-mean-field}. The final equations are
presented in the main text.

\section{Solutions to the dynamic mean-field equations with specific priors}

In order to theoretically evaluate the performance of the AMP algorithms
given different tensors, we explicitly derive the dynamic mean-field
equation of the overlap, and the error, for some common priors $P_{\alpha}(x)$.
These derivations closely resemble the analysis done in \cite{lesieur2017constrained}
for the $p=2$ case, only here we allow different mixture of prior
and different sizes for the modes, and consider arbitrary order $p$.
In the following we will use rank $r=1$ tensors, where the estimators
$x$, the ground truth $x^{0}$, and the the overlaps $M$ , which
we will denote here as $m$, are all scalar values. The same analysis
hods with multivariate calculation, when $r\geq2$.

In order to theoretically evaluate the performance of the AMP algorithms
given different tensors, we explicitly derive the dynamic mean-field
equations of the overlap, and the error, for some common priors $P_{\alpha}(x)$.
These derivations closely resemble the analysis done in \cite{lesieur2017constrained}
for the $p=2$ case, only here we allow a different mixture of prior
and different sizes for the modes, and consider arbitrary order $p$.
In the following, we will use rank $r=1$ tensors, where the estimators
$x$, the ground truth $x^{0}$, and the overlaps $M$ -- which we
will denote here as m -- are all scalar values. The same analysis
holds with multivariate calculation when $r\geq2$.

\subsection{Gaussian prior}

The first, and perhaps most common, choice for prior is a normal distribution
of $x_{\alpha}$ , with variance $\sigma_{\alpha}^{2}$ and and mean
$\mu_{\alpha}$,
\begin{equation}
P_{\alpha}(x)=\frac{1}{\sqrt{2\pi}\sigma_{\alpha}}e^{-(x-\mu_{\alpha})^{2}/2\sigma_{\alpha}^{2}}.
\end{equation}
We use that prior to explicitly calculate the update rule, 
\begin{equation}
f_{\alpha}=\frac{\partial}{\partial\mathbf{u}}\Z_{\alpha}(A,\mathbf{u})=\frac{\intop d\mathbf{x}P_{\alpha}(\mathbf{x})\mathbf{x}e^{-\frac{1}{2}\mathbf{x}^{T}A\mathbf{x}+\mathbf{u}^{T}\mathbf{x}}}{\intop d\mathbf{x}P_{\alpha}(\mathbf{x})e^{-\frac{1}{2}\mathbf{x}^{T}A\mathbf{x}+\mathbf{u}^{T}\mathbf{x}}}.\label{eq:f}
\end{equation}
The nominator of \eqref{eq:f} can be written as
\begin{multline}
\intop dxx\frac{1}{\sqrt{2\pi}}e^{-(x-\mu_{\alpha})^{2}/2\sigma_{\alpha}^{2}}e^{-\frac{1}{2}Ax^{2}+ux}\\
=\intop dxx\frac{1}{\sqrt{2\pi}\sigma_{\alpha}}\exp\frac{1}{2\sigma_{\alpha}^{2}}\left[-x^{2}+2x\mu_{\alpha}-\mu_{\alpha}^{2}-ax+2bx\right]\\
=\intop dxx\frac{1}{\sqrt{2\pi}\sigma_{\alpha}}\exp\frac{-1}{2\sigma_{\alpha}^{2}}\left[(a+1)x^{2}-2(\mu_{\alpha}+b)x+\mu_{\alpha}^{2}\right],
\end{multline}
where $b=u\sigma_{\alpha}^{2}$ and $a=A\sigma_{\alpha}^{2}$.Completing
the quadratic form, we have
\begin{multline*}
=\intop dxx\frac{1}{\sqrt{2\pi}\sigma_{\alpha}}\exp\frac{-1}{2\sigma_{\alpha}^{2}}\left[\left(\sqrt{a+1}x-\frac{\mu_{\alpha}+b}{\sqrt{a+1}}\right)^{2}+\mu_{\alpha}^{2}-\frac{(\mu_{\alpha}+b)^{2}}{a+1}\right]\\
=\frac{1}{\sqrt{a+1}}\exp\left[\frac{-1}{\sigma_{\alpha}^{2}}\mu_{\alpha}^{2}-\frac{(\mu_{\alpha}+b)^{2}}{a+1}\right]\intop dxx\frac{\sqrt{a+1}}{\sqrt{2\pi}\sigma_{\alpha}}\exp\frac{-(a+1)}{\sigma_{\alpha}^{2}}\left[\left(x-\frac{\mu_{\alpha}+b}{a+1}\right)^{2}\right].
\end{multline*}
Similar treatment in performed on the denominator. It is straight
forward to see that the function in \eqref{eq:f} reduces to
\begin{equation}
f_{\alpha}(A,u)=\frac{\mu_{\alpha}+u\sigma_{\alpha}^{2}}{A\sigma_{\alpha}^{2}+1}.
\end{equation}
Next, we want to use this functional form in the dynamic mean-field
eq. \eqref{eq:DMFT_SE-1}. Denote $\tilde{m}_{\alpha}^{t}=\frac{1}{\Delta}\prod_{\beta\neq\alpha}^{\odot}m_{\beta}^{t}$,
then we have $A^{t}=\hat{m}_{\alpha}^{t}$ and $\mathbf{u}_{\alpha}^{t}=\hat{m}_{\alpha}^{t}x_{\alpha}^{0}+\sqrt{\hat{m}_{\alpha}^{t}}z$
then we want to compute 
\begin{equation}
\left\langle \frac{\frac{\mu_{\alpha}}{\sigma_{\alpha}^{2}}+\mathbf{u}}{A+\frac{1}{\sigma_{\alpha}^{2}}}x^{0}\right\rangle _{z,x_{\alpha}^{0}}=\left\langle \frac{\frac{\mu_{\alpha}}{\sigma_{\alpha}^{2}}+\hat{m}_{\alpha}^{t}x_{\alpha}^{0}+\sqrt{\tilde{m}_{\alpha}^{t}}z}{\tilde{m}_{\alpha}^{t}+\frac{1}{\sigma_{\alpha}^{2}}}x^{0}\right\rangle _{z,x_{\alpha}^{0}}=\left\langle \frac{\frac{\mu_{\alpha}}{\sigma_{\alpha}^{2}}+\tilde{m}_{\alpha}^{t}x_{\alpha}^{0}}{\tilde{m}_{\alpha}^{t}+\frac{1}{\sigma_{\alpha}^{2}}}x^{0}\right\rangle _{x^{0}},
\end{equation}
Averaging over the distribution of the ground-truth values $P(x^{0})$,
\begin{equation}
\left\langle f_{\alpha}\left(\tilde{m}_{\alpha}^{t}\right)x^{0}\right\rangle _{P_{\alpha}}=\frac{\frac{\mu_{\alpha}^{2}}{\sigma_{\alpha}^{2}}+\tilde{m}_{\alpha}^{t}\left(\sigma_{\alpha}^{2}+\mu_{\alpha}^{2}\right)}{\tilde{m}_{\alpha}^{t}+\frac{1}{\sigma_{\alpha}^{2}}}.
\end{equation}
Finally the dynamic mean-field iterative equation on the mean overlap
are given by 
\begin{equation}
m_{\alpha}^{t+1}=\frac{\Delta\frac{\mu_{\alpha}^{2}}{\sigma_{\alpha}^{2}}+\left(\sigma_{\alpha}^{2}+\mu_{\alpha}^{2}\right)\prod_{\beta\neq\alpha}^{\odot}m_{\beta}^{t}}{\frac{\Delta}{\sigma_{\alpha}^{2}}+\prod_{\beta\neq\alpha}^{\odot}m_{\beta}^{t}}.\label{eq:DMFT_gauss-1}
\end{equation}
In the case of zero mean priors, $\mu_{\alpha}=0$, the equation is
reduced to
\begin{equation}
m_{\alpha}^{t+1}=\frac{\sigma_{\alpha}^{2}\prod_{\beta\neq\alpha}^{\odot}m_{\beta}^{t}}{\frac{\Delta}{\sigma_{\alpha}^{2}}+\prod_{\beta\neq\alpha}^{\odot}m_{\beta}^{t}}.
\end{equation}
We note that if all modes are Gaussian with zero means, then the solution
$M_{\alpha}=0$ $\forall\alpha$ is a stable fixed point of the dynamics,
implying that if we start from random initial conditions, that are
uncorrelated with the true values, the algorithms will not converge.
A numerical analysis of eq. \eqref{eq:DMFT_gauss-1} for order $p=3$
tensors is presented in the main text.

Consider the case of $\mu_{\alpha}=\mu$ and $\sigma_{\alpha}=\sigma$,
with all priors are similar. From the structure of \eqref{eq:DMFT_gauss-1}
we find that in the fixed point $M_{\alpha}^{*}$
\begin{equation}
m_{\alpha}^{*}=m^{*}\;\forall\alpha,
\end{equation}
which is what would be expected from the symmetry of the problem.
Note however that unlike the derivation in \cite{lesieur2017statistical},
the underlying tensor in non-symmetric. 

\textbf{Noncubic tensors}. If we have different population sizes,
then we have a ratio between the order parameters, and replace $m_{\alpha}$with
$n_{\alpha}m_{\alpha}$. 
\begin{equation}
m_{\alpha}^{t+1}=\frac{\Delta\frac{\mu_{\alpha}^{2}}{\sigma_{\alpha}^{2}}+\left(\sigma_{\alpha}^{2}+\mu_{\alpha}^{2}\right)\prod_{\beta\neq\alpha}n_{\beta}m_{\beta}^{t}}{\frac{\Delta}{\sigma_{\alpha}^{2}}+\prod_{\beta\neq\alpha}k_{\beta}m_{\beta}^{t}}
\end{equation}

\subsection{Bernoulli distribution}

For many applications, it is expected that some of the modes in the
underlying low rank tensors are sparse, meaning they contribute information
to only a small subset of the measurements. A simple way of modeling
such data is using the Bernoulli distribution, 
\begin{equation}
P_{\alpha}(x)=\rho\delta(x-1)+(1-\rho)\delta(x).
\end{equation}

As in the derivation of the Gaussian priors in the previous subsection,
we compute the function \eqref{eq:f}. The nominator is equal to 
\begin{multline}
\intop dx\left[\rho\delta(x-1)+(1-\rho)\delta(x)\right]xe^{-\frac{1}{2}x^{T}Ax+\mathbf{u}^{T}x}=\\
\rho e^{-\frac{1}{2}\sum_{ij}A+\sum_{j}\mathbf{u_{j}}},
\end{multline}
and the denominator is given by
\begin{multline}
\intop dx\left[\rho\delta(x-1)+(1-\rho)\delta(x)\right]e^{-\frac{1}{2}x^{T}Ax+\mathbf{u}^{T}x}=\\
\rho e^{-\frac{1}{2}\sum_{ij}A+\sum_{j}\mathbf{u}}+(1-\rho).
\end{multline}
Combining both expressions together we get
\begin{equation}
f_{\alpha}(A,u)=\frac{\rho e^{-\frac{1}{2}A+\mathbf{u}}}{\rho e^{-\frac{1}{2}A+\mathbf{u}}+(1-\rho)}=\frac{\rho}{\rho+(1-\rho)e^{\frac{1}{2}A-\mathbf{u}}},
\end{equation}
with first derivative equal to
\begin{equation}
\frac{\partial}{\partial\mathbf{u}}f_{\alpha}=\frac{e^{-\frac{1}{2}A+\mathbf{u}}\left(\rho^{-1}-1\right)}{\left[\left(e^{-\frac{1}{2}A+\mathbf{u}}-1\right)+\rho^{-1}\right]^{2}}.\label{eq:Bernouli_f}
\end{equation}

In the bayes optimal case we have $A_{\alpha}=\tilde{m}_{\alpha}$
and $u_{\alpha i}=\tilde{m}_{\alpha}x_{i}^{0}+\sqrt{\tilde{m}_{\alpha}}z$
,where we have defined $\tilde{m}_{\alpha}\equiv\frac{1}{\Delta}\prod_{\beta\neq\alpha}m_{\beta}$.
In the expression in the exponent of \eqref{eq:Bernouli_f} we have
\begin{equation}
\frac{1}{2}A-u_{i}=\tilde{m}_{\alpha}\left(\frac{1}{2}-x_{i}^{0}\right)+\sqrt{\tilde{m}_{\alpha}}z.
\end{equation}
Next we integrate over the prior and ground-truth to get 
\begin{multline}
m_{\alpha}^{t+1}=\rho\E_{z}\left[f_{\alpha}\left(\tilde{m}_{\alpha},\tilde{m}_{\alpha}+\sqrt{\tilde{m}_{\alpha}}z\right)\right]=\\
\rho^{2}\left\langle \left(\rho+(1-\rho)\exp\left[\frac{1}{2}\tilde{m}_{\alpha}-\sqrt{\tilde{m}_{\alpha}}z\right]\right)^{-1}\right\rangle _{z}
\end{multline}
In the sparse case, where $\rho\ll1$this can be simplified further
\begin{multline}
m_{\alpha}^{t+1}=\rho^{2}\left\langle \left(\exp\left[-\frac{1}{2}\tilde{m}_{\alpha}+\sqrt{\tilde{m}_{\alpha}}z\right]\right)\right\rangle _{z}+\mathcal{O}(\rho^{3})\\
=\frac{\rho^{2}}{\sqrt{2\pi}}\intop_{-\infty}^{\infty}dz\exp\left[-\frac{z^{2}}{2}-\frac{1}{2}\tilde{m}_{\alpha}+\sqrt{\tilde{m}_{\alpha}}z\right]+\mathcal{O}(\rho^{3})\\
=\rho^{2}e^{\tilde{m}_{\alpha}/2}+\mathcal{O}(\rho^{3})
\end{multline}
Note that in a complete overlap we have $m=\rho$ so $e^{\tilde{m}_{\alpha}/2}=1/\rho$
and 
\[
\frac{1}{2\Delta}\prod_{\beta\neq\alpha}m_{\beta}=-\log\rho
\]
In instances where all of the modes have similar statistics, then
we would have 
\[
\frac{1}{2\Delta}\rho^{p-1}=-\log\rho\Rightarrow\Delta=\frac{\rho^{p-1}}{2\left|\log\rho\right|}.
\]
Here, we can expect that for $\Delta\sim\rho^{p-1}/\left|\log\rho\right|$,
where $p$ is the order of the tensor, we will have high overlap with
zero error. However, in the case of non-symmetric tensors, not all
directions have to be sparse, and may have different distributions.
In that case the noise scale as $\Delta\sim\rho^{\tilde{p}-1}/\left|\log\rho\right|$,
where $\tilde{p}$ is the number of sparse modes in the underlying
tensor. 

\subsection{Gauss-Bernoulli}

The next logical step is to combine the continuous irregularity of
the Gaussian distribution and the sparse nature of the Bernoulli distribution.
The Gauss-Bernoulli distribution is given by

\begin{equation}
P_{\alpha}(x)=\rho\mathcal{N}(\mu,\sigma^{2})+(1-\rho)\delta(x).
\end{equation}
For brevity we will use zero mean $\mu=0$ and unit variance $\sigma^{2}=1$,
and note that the results can be easily rescaled. The update function
is given by
\[
f_{\alpha}(A,u)=\frac{\rho\intop dxx\frac{1}{\sqrt{2\pi}}e^{-\frac{1}{2}x^{T}Ax+\mathbf{u}^{T}x-\frac{1}{2}x^{2}}}{\rho\intop dx\frac{1}{\sqrt{2\pi}}e^{-\frac{1}{2}x^{T}Ax+\mathbf{u}^{T}x-\frac{1}{2}x^{2}}+(1-\rho)}.
\]
Using some algebra we get
\begin{multline}
f_{\alpha}=\frac{\rho\intop dxx\frac{1}{\sqrt{2\pi}}\exp\left[-\frac{1}{2}(\sqrt{A-1}x-\frac{u}{\sqrt{A+1}})^{2}+\frac{u^{2}}{(A+1)}\right]}{\rho\intop dx\frac{1}{\sqrt{2\pi}}\exp\left[-\frac{1}{2}(\sqrt{A-1}x-\frac{u}{\sqrt{A+1}})^{2}+\frac{u^{2}}{(A+1)}\right]+(1-\rho)}=\\
\frac{\frac{1}{\sqrt{A+1}}\rho\intop dxx\frac{\sqrt{A+1}}{\sqrt{2\pi}}\exp\left[-\frac{\sqrt{A-1}}{2}(x-\frac{u}{A+1})^{2}\right]e^{\frac{u^{2}}{(A+1)}}}{\frac{1}{\sqrt{A+1}}\rho\intop dx\frac{\sqrt{A+1}}{\sqrt{2\pi}}\exp\left[-\frac{\sqrt{A-1}}{2}(x-\frac{u}{A+1})^{2}\right]e^{\frac{u^{2}}{(A+1)}}+(1-\rho)}=\\
\frac{\rho u}{(A+1)\rho+(1-\rho)\left(A+1\right)^{3/2}e^{\frac{-u^{2}}{(A+1)}}}
\end{multline}
and the first derivative is given by
\begin{equation}
\frac{\partial f_{\alpha}}{\partial u}=\rho\frac{\left((A+1)\rho+(1-\rho)\left(A+1\right)^{3/2}e^{\frac{-u^{2}}{(A+1)}}\right)+2(1-\rho)u^{2}\left(A+1\right)^{1/2}e^{\frac{-u^{2}}{(A+1)}}}{\left((A+1)\rho+(1-\rho)\left(A+1\right)^{3/2}e^{\frac{-u^{2}}{(A+1)}}\right)^{2}}.
\end{equation}

For sanity check, if $\rho=1$ then 
\[
f_{\alpha}(\rho=1)=\frac{u}{A+1}
\]
\[
\frac{\partial f_{\alpha}\left(\rho=1\right)}{\partial u}=\frac{1}{A+1}
\]
and we have recovered the results for the Gaussian priors from above.
From here we can calculate the dynamic mean-filed equations 
\begin{equation}
m_{\alpha}^{t+1}=\rho\int P_{\alpha}(x^{0})dx^{0}\frac{dz}{\sqrt{2\pi}}e^{-z^{2}/2}\frac{\tilde{m}_{\alpha}^{t}x^{0}+\sqrt{\frac{1}{\Delta}\tilde{m}_{\alpha}^{t}}z}{(\tilde{m}_{\alpha}^{t}+1)\rho+(1-\rho)\left(\tilde{m}_{\alpha}^{t}+1\right)^{3/2}\exp\left[-\frac{\left(\tilde{m}_{\alpha}^{t}x^{0}+\sqrt{\tilde{m}_{\alpha}^{t}}z\right)^{2}}{\tilde{m}_{\alpha}^{t}+1}\right]}x^{0}
\end{equation}

\[
=\rho^{2}\frac{\tilde{m}_{\alpha}^{t}}{(\tilde{m}_{\alpha}^{t}+1)}\int\frac{dzdx^{0}}{2\pi}\exp\left(-\frac{x^{02}+z^{2}}{2}\right)\frac{\left(x^{0}\right)^{2}}{\rho+(1-\rho)\left(\tilde{m}_{\alpha}^{t}+1\right)^{1/2}\exp\left[-\frac{\left(\tilde{m}_{\alpha}^{t}x^{0}+\sqrt{\tilde{m}_{\alpha}^{t}}z\right)^{2}}{\tilde{m}_{\alpha}^{t}+1}\right]}
\]

\subsection{Mixed priors}

In the case of general asymmetric tensors, we can construct a tensor
using different priors for the different modes. It is particularly
useful in real applications, as different modes of the tensors can
originate from entirely different sources. Consider for example an
order-3 tensor holding neural firing rate data $r_{itk}$. The index
$i$ marks the neuron recorded; index $t$ is the time bin within
a single trial, and $k$ is the trial index. If we believe that the
data originates from the low-dimensional dynamical system, we would
want to write the tensor as

\begin{equation}
r_{itk}=\sum_{\rho}^{D}u_{i}^{\rho}x_{t}^{\rho}v_{k}^{\rho}+\sqrt{\Delta}\epsilon_{itk},
\end{equation}
where $D$ is the dimensions of the dynamical system, and $\Delta$
is the noise of a single measurement.We may ask how should we design
an experiment so that low-rank decomposition of the recorded data
would be possible. In this case, we would assert different priors
to the different modes. The mode $x_{t}$ represent the $D$ dimensional
dynamical system. We could assume for example that is generated by
some Gaussian process, thus follows Gaussian statistics. The mode
$u_{i}$ represents the projections of the low-dimensional dynamical
system onto the set measured neurons. It may be a valid assumption
that only a fraction of the neurons responds in coherence with the
underlying dynamics; a Gauss-Bernoulli distribution will be suitable
for this mode. Lastly, the trial modulus mode $v_{k}$ can have Gaussian
distribution about some mean with small variance, suggesting small
trial-to-trial modulations.

To solve the dynamic mean field theory for this case, and find the
boundaries of the inference we would use the appropriate equation
for each of the modes. For example, for two Gaussian distributions
and one Gauss-Bernoulli, we would have
\begin{align}
m_{x}^{t+1} & =\frac{\Delta\frac{\mu_{x}^{2}}{\sigma_{x}^{2}}+\left(\sigma_{x}^{2}+\mu_{x}^{2}\right)m_{u}^{t}m_{v}^{t}}{\frac{\Delta}{\sigma_{\alpha}^{2}}+m_{u}^{t}m_{v}^{t}}\\
m_{v}^{t+1} & =\frac{\Delta\frac{\mu_{v}^{2}}{\sigma_{v}^{2}}+\left(\sigma_{v}^{2}+\mu_{v}^{2}\right)m_{u}^{t}m_{x}^{t}}{\frac{\Delta}{\sigma_{\alpha}^{2}}+m_{u}^{t}m_{x}^{t}}\\
m_{u}^{t+1} & =\rho^{2}\frac{\tilde{m}_{\alpha}^{t}}{(\tilde{m}_{\alpha}^{t}+1)}\int\frac{dzdx^{0}}{2\pi}\exp\left(-\frac{x^{02}+z^{2}}{2}\right)\\
 & \times\frac{\left(x^{0}\right)^{2}}{\rho+(1-\rho)\left(\frac{1}{\Delta}m_{x}^{t}m_{v}^{t}+1\right)^{1/2}\exp\left[-\frac{\left(\frac{1}{\Delta}m_{x}^{t}m_{v}^{t}x^{0}+\sqrt{\frac{1}{\Delta}m_{x}^{t}m_{v}^{t}}z\right)^{2}}{\frac{1}{\Delta}m_{x}^{t}m_{v}^{t}+1}\right]}
\end{align}

This set of equations can be solved numerically, to find an estimate
for AMP performances under the noise. 

\bibliographystyle{unsrt}
\bibliography{TensorDecomp}

\end{document}